\documentclass[10pt,twocolumn,letterpaper]{article}

\usepackage[pagenumbers]{cvpr}

\usepackage{xcolor}
\usepackage{multirow}

\definecolor{cvprblue}{rgb}{0.21,0.49,0.74}
\usepackage[pagebackref,breaklinks,colorlinks,allcolors=cvprblue]{hyperref}

\newcommand{\muted}[1]{{%
    \hypersetup{citecolor=black!55}%
    \color{black!55}#1%
}}

\def\confName{CVPR}
\def\confYear{2027}

\title{LandV: Multimodal World Modeling for Autonomous Driving via Modality-Aligned Asymmetric Dual-Stream Architecture}

\author{%
  Jiasong Feng$^{1,2}$\quad Chen Liu$^{2\dagger}$\quad Zhanjie Zhang$^{3}$\quad Yimeng Li$^{2}$\\
  Mofan Zhou$^{2}$\quad Kun Zhan$^{2}$\quad Qilong Wang$^{1*}$\\[4pt]
  $^{1}$Tianjin University\quad $^{2}$Li Auto Inc.\quad $^{3}$Zhejiang University\\[2pt]
  {\normalsize
    \href{https://landv-wm.github.io/}{\raisebox{-.12em}{\includegraphics[height=.9em]{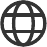}}\hspace{.28em}Project Page}\qquad
    \href{https://github.com/fjscfy/LandV}{\raisebox{-.12em}{\includegraphics[height=.9em]{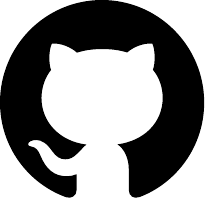}}\hspace{.28em}GitHub}\qquad
    \href{https://huggingface.co/feng8/LandV}{\raisebox{-.12em}{\includegraphics[height=.9em]{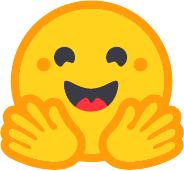}}\hspace{.28em}Hugging Face}}
}
\hypersetup{pdfauthor={Jiasong Feng, Chen Liu, Zhanjie Zhang, Yimeng Li, Mofan Zhou, Kun Zhan, Qilong Wang}}

\begin{document}
\maketitle
\begingroup
  \renewcommand{\thefootnote}{\fnsymbol{footnote}}
  \footnotetext[2]{Project Lead.\quad $^{*}$Corresponding Author.}
\endgroup
\begin{abstract}
Joint video--LiDAR generation is essential for multimodal driving world models to simulate dynamic scenes with consistent visual and geometric observations. However, existing approaches either compromise modality-specific modeling through shared generative backbones or limit cross-modal interaction by relying on intermediate scene representations. We propose \textbf{LandV}, an asymmetric dual-stream multimodal world model for autonomous driving that preserves modality-specific \textbf{L}iDAR \textbf{and} \textbf{V}ideo generation while enabling effective cross-modal interaction through explicit constraints on geometric correspondences.
Specifically, our LandV employs dedicated generative streams tailored to the characteristics of each modality, with a lightweight, VAE-free LiDAR branch and a Local Geometry Refinement (LGR) head for fine-grained geometric modeling. To effectively coordinate the two streams, our LandV presents geometry-guided bidirectional cross-attention for efficient cross-modal interaction and an Adaptive Reciprocal Alignment (ARA) loss to ensure cross-modal consistency.
A unified benchmark is established on nuScenes covering generation quality, video object consistency, and cross-modal geometric alignment for extensive evaluation, where the results  demonstrate that our LandV achieves very competitive generation quality for both modalities while substantially improving cross-modal geometric consistency.

\end{abstract}
    
\section{Introduction}
\label{sec:intro}

Driving world models have made significant progress~\cite{gaia1,drivewm,vista,drivingworld,opendwm}. As autonomous driving systems rely on multiple sensors to perceive their surroundings~\cite{bevfusion}, multimodal driving world models that jointly generate videos and LiDAR scans have attracted increasing attention~\cite{uniscene,uniscenev2,unidrivedreamer,omnigen,m4world,genesis}. Compared with methods that generate videos~\cite{drivedreamer,drivingdiffusion,magicdrivev2,drivedreamer2} or LiDAR scans~\cite{lidargen,r2dm,rangeldm,lagen} independently, joint generation holds the potential to produce cross-modally consistent observations, thereby better supporting downstream tasks. Moreover, leveraging the complementary visual and geometric information of the two modalities could improve generation quality for both.

\begin{figure}[t]
	\centering
	\includegraphics[width=\linewidth]{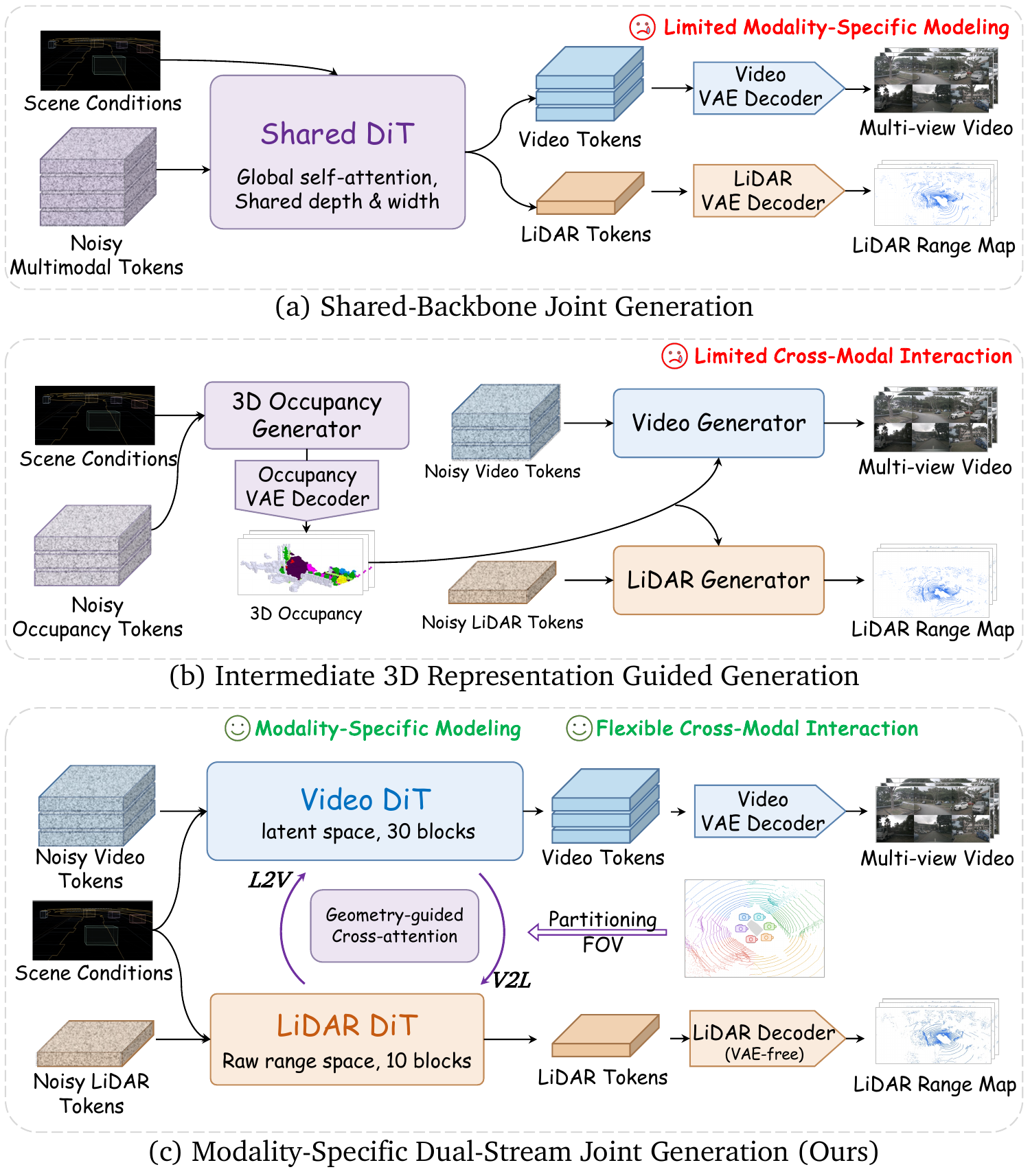}
	\caption{\textbf{Comparison of joint video-LiDAR generation paradigms.}
	(a) Shared-backbone methods~\cite{unidrivedreamer,omnigen,m4world} jointly model video and LiDAR using a unified backbone.
	(b) Intermediate-representation methods~\cite{uniscene,uniscenev2,genesis} generate both modalities from a common 3D representation.
	(c) Our LandV adopts asymmetric modality-specific branches with geometry-guided cross-attention for efficient cross-modal interaction.}
	\label{fig:paradigm-comparison}
\end{figure}

A core challenge in joint video–LiDAR is how to simultaneously optimize the generation processes of the two modalities effectively. Existing methods can be roughly categorized into two classes, i.e., shared-backbone methods~\cite{unidrivedreamer,omnigen,m4world} and intermediate-representation methods ~\cite{uniscene,uniscenev2,genesis}. As shown in Fig.~\ref{fig:paradigm-comparison}(a), shared-backbone methods~\cite{unidrivedreamer,omnigen} map video and LiDAR into a unified latent space and jointly model them using a shared backbone. Intermediate-representation methods~\cite{uniscene,uniscenev2,genesis} employ an intermediate 3D scene representation (e.g., occupancy), to provide a common geometric basis for generating both modalities, as shown in Fig.~\ref{fig:paradigm-comparison}(b).
Despite these advances, both shared-backbone and intermediate-representation methods still struggle to balance modality-specific modeling and cross-modal interaction. Specifically, representation learning based on a unified backbone generally suffers from the cross-modal interference, hindering the effective modeling of modality-specific characteristics~\cite{mot,unix,alignmif}. Meanwhile, intermediate-representation methods~\cite{uniscene,uniscenev2,genesis} employ the common intermediate features to generate video and LiDAR through two separate branches, where generation process overlooks  effective cross-modal interaction. Particularly, UniScene~\cite{uniscene} with ground-truth intermediate occupancy only brings 2.79\% improvement (61.22\% to 64.01\%) and is far below the ground-truth score of 80.36\% (see Tab.~\ref{tab:generation-quality} for details), indicating the consistency gap between video and LiDAR generation hardly be mitigated by precise occupancy generation.

To address these limitations, we propose \textbf{LandV}, an asymmetric dual-stream multimodal world model for autonomous driving, which aims to preserve modality-specific \textbf{L}iDAR \textbf{and} \textbf{V}ideo generation while enabling effective cross-modal interaction through explicit constraints on geometric correspondences, as shown in Fig.~\ref{fig:paradigm-comparison}(c). For modality-specific modeling, our LandV proposes an asymmetric dual-stream architecture with a Local Geometry Refinement (LGR) head to account for the characteristics of video and LiDAR. Specifically, the video stream uses pretrained VAE latents~\cite{wan} and v-prediction, while the LiDAR stream specializes a lightweight backbone to directly generate range maps with x-prediction~\cite{li2026back}, avoiding training an intractable LiDAR VAE~\cite{sgldm,ultralidar}. Furthermore, the LGR head integrates backbone features with local information from noisy LiDAR inputs to compensate for fine-grained geometric details in VAE-free LiDAR generation.
For effective cross-modal interaction, our LandV presents geometry-guided bidirectional cross-attention with an Adaptive Reciprocal Alignment (ARA) to enable efficient interaction and object-level alignment between video and LiDAR features. Specifically, a sensor calibration is first exploited to restrict attention into spatially corresponding regions for reducing computational cost. Then, our ARA establishes object-level geometric correspondences using 3D annotations and LiDAR-to-image projections. It selects the more object-aligned attention distribution as the teacher to supervise the other through KL divergence, encouraging cross-modal consistency.

Extensive experiments on nuScenes demonstrate the effectiveness of LandV.
Particularly, under the defined metrics of geometric consistency, our proposed LandV clearly surpasses UniScene~\cite{uniscene} even with ground-truth occupancy, while approaching the ground-truth score. Our contributions are summarized as follows:
\begin{itemize}
    \item We propose \textbf{LandV}, a first asymmetric dual-stream architecture for effective joint video-LiDAR generation for autonomous driving, which takes into account both modality-specific representation learning  and effective cross-modal interaction.

    \item To this end, our LandV presents modality-specific backbones for video and LiDAR generation with geometry-guided bidirectional cross-attention for video and LiDAR representation learning. Then, geometry-guided bidirectional cross-attention is developed for efficient yet effective cross-modal interaction, along with an Adaptive Reciprocal Alignment (ARA) loss on object-level geometric correspondences for ensuring cross-modal consistency.

    \item Extensive experiments on a unified benchmark for video–LiDAR generation demonstrate our LandV achieves very competitive generation quality for both modalities, while showing better cross-modal consistency under the established metrics in terms of video object consistency and cross-modal geometric alignment. 
\end{itemize}

\begin{figure*}[t]
    \centering
    \includegraphics[width=\textwidth]{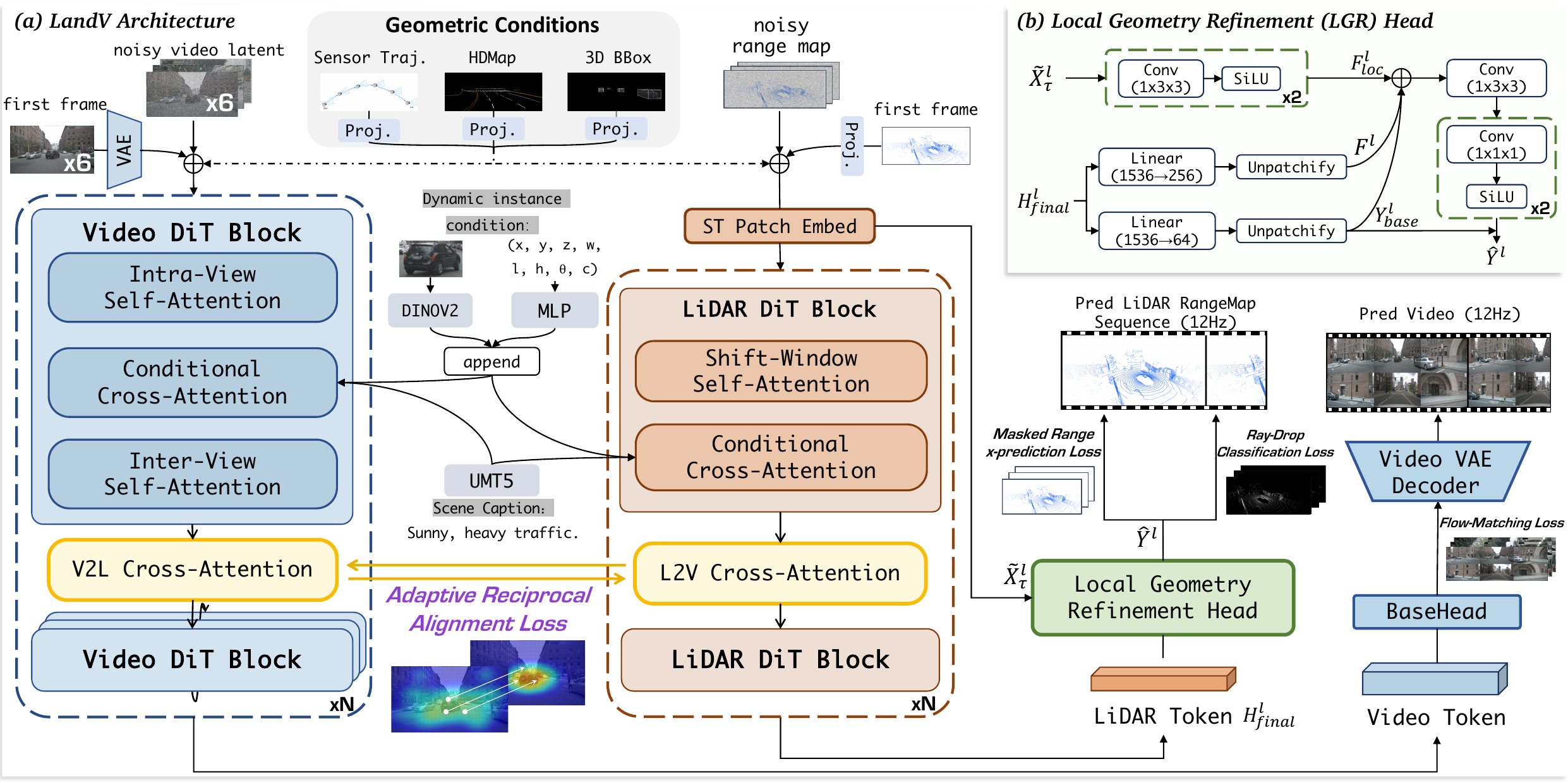}
    \caption{\textbf{Overview of our LandV.} (a) our LandV proposes an asymmetric dual-stream architecture for joint video--LiDAR generation. Specifically, the video stream operates in a pretrained VAE latent space, while the LiDAR stream directly models range maps with a lightweight DiT. Both streams incorporate scene and geometric conditions, and perform cross-modal interaction by our geometry-guided bidirectional cross-attention, being further constrained by the Adaptive Reciprocal Alignment (ARA) loss. (b) For LiDAR generation, our Local Geometry Refinement (LGR) head integrates 2-channel base predictions $Y_{\mathrm{base}}^l$ as LiDAR estimates, 8-channel backbone features $F^l$ retaining rich contextual information from the DiT, and 16-channel local features $F_{\mathrm{loc}}^l$ accounting for fine-grained spatial structures from noisy LiDAR inputs. These features are fused to predict a residual correction to $Y_{\mathrm{base}}^l$, yielding refined 2-channel LiDAR predictions $\hat Y^l$.}
    \label{fig:landv-framework}
\end{figure*}

\section{Related Work}
\label{sec:related_work}

\subsection{World Models for Autonomous Driving}

Driving world models support simulation and planning by modeling future scenes through different representations, including video, LiDAR scans, semantic occupancy, and others.
For video generation, GAIA-1~\cite{gaia1} incorporates multimodal conditioning, DrivingWorld~\cite{drivingworld} enables autoregressive long-horizon prediction, while Drive-WM~\cite{drivewm} and Vista~\cite{vista} explore planning and action controllability.
MagicDrive~\cite{magicdrive} and MagicDrive-V2~\cite{magicdrivev2} introduce geometric and semantic conditions for multi-view generation.
For LiDAR, R2DM~\cite{r2dm} and RangeLDM~\cite{rangeldm} adopt diffusion in range-image and latent spaces, respectively, while LaGen~\cite{lagen} generates autoregressive LiDAR sequences.
For occupancy, OccWorld~\cite{occworld} predicts future semantic occupancy to model scene evolution.
However, these methods primarily focus on individual scene representations, limiting their ability to provide cross-modally consistent visual and geometric observations for autonomous driving tasks.

\subsection{Joint video and LiDAR Generation}

Existing approaches to joint video and LiDAR generation can be broadly categorized into shared-backbone and intermediate-representation methods.
For shared-backbone methods, OmniGen~\cite{omnigen} employs a unified model for joint image and LiDAR synthesis using a shared BEV representation.
UniDriveDreamer~\cite{unidrivedreamer} and M$^4$World~\cite{m4world} generate video and LiDAR sequences by mapping both modalities into a unified VAE latent space and modeling them with a shared DiT backbone.
However, shared-backbone methods restrict modality-specific modeling and may introduce cross-modal interference.
For intermediate-representation methods, Genesis~\cite{genesis} adopts a two-stage pipeline combining video diffusion with BEV-aware LiDAR generation through NeRF-based rendering. UniScene~\cite{uniscene} employs semantic occupancy to guide video and LiDAR generation.
UniSceneV2~\cite{uniscenev2} extends this framework with occupancy forecasting and sensor-specific generation. However, these methods generate both modalities through separate branches with limited cross-modal interaction, making it difficult to ensure cross-modal consistency.

\section{Method}
\label{sec:method}

As illustrated in Fig.~\ref{fig:landv-framework}(a), our LandV jointly generates multi-view videos and LiDAR sequences through an asymmetric dual-stream architecture.
Specifically, given initial observations and scene conditions, the video stream operates in a pretrained VAE latent space with intra-view and inter-view attention, while the lightweight LiDAR stream directly processes range maps using spatiotemporal patch embedding and shifted-window self-attention. 
To enable efficient cross-modal interaction, geometry-guided bidirectional cross-attention connects the two streams.
Furthermore, our LandV trains each stream with a modality-specific prediction objective to accommodate the distinct characteristics of video and LiDAR.
The conditioning, architecture, and training objectives are detailed in Sec.~\ref{sec:joint-generation}.
To enhance fine-grained geometric fidelity in LiDAR generation, we introduce a Local Geometry Refinement (LGR) head that integrates backbone features with local information from noisy range maps, as illustrated in Fig.~\ref{fig:landv-framework}(b) and detailed in Sec.~\ref{sec:local-refinement}.
For cross-modal consistency, we further introduce an Adaptive Reciprocal Alignment (ARA) loss that leverages object-level geometric correspondences to adaptively align bidirectional cross-attention distributions, as illustrated in Fig.~\ref{fig:ara} and detailed in Sec.~\ref{sec:attention-alignment}.

\subsection{Asymmetric Dual-Stream Architecture}
\label{sec:joint-generation}

\noindent\textbf{Scene Conditioning.}
Our LandV conditions both streams on scene semantics, geometry, and initial observations, with modality-specific representations adapted to video and LiDAR.
Specifically, these conditions are organized into spatial and token-based representations.
Our LandV adopts spatial conditioning for scene layout, sensor geometry, and initial observations, as they can be explicitly aligned with the spatial grids of each modality. Using sensor calibration, HD maps and 3D bounding boxes are mapped to modality-specific spatial grids to construct $C_{\mathrm{map}}^m$ and $C_{\mathrm{box}}^m$, respectively. Here, $m\in\{v,l\}$ denotes the video or LiDAR modality.
The sensor geometry itself is represented by Pl\"ucker ray maps $C_{\mathrm{ray}}^m$, while $C_{\mathrm{init}}^m$ represents the first-frame observation of each modality.
For token-based conditioning, we encode scene captions and dynamic-object attributes as discrete tokens. Scene captions are encoded into text embeddings $C_{\mathrm{text}}^m$ using UMT5~\cite{umt5}. Dynamic-object attributes include 3D bounding-box parameters $(x,y,z,w,l,h,\theta)$, object category $c$, and reference-image appearance features extracted by DINOv2~\cite{dinov2}. These attributes are separately encoded using dedicated linear layers to construct object tokens $C_{\mathrm{obj}}^m$. Token-based conditions are incorporated through cross-attention, while spatial conditions are concatenated and projected onto the corresponding input token features:
\begin{equation}
\label{eq:condition-input}
H_0^m = E_{\mathrm{in}}^m(X_\tau^m)
+ P_{\mathrm{sp}}^m\left(
[C_{\mathrm{map}}^m; C_{\mathrm{box}}^m;
C_{\mathrm{ray}}^m; C_{\mathrm{init}}^m]
\right),
\end{equation}
where $E_{\mathrm{in}}^m$ denotes the modality-specific patch embedding, $P_{\mathrm{sp}}^m$ is a channel projection convolution, and $[\,;\,]$ denotes channel-wise concatenation.

\noindent\textbf{Modality-Specific Modeling.} Our LandV designs the two streams with distinct representations and attention mechanisms, enabling each to perform conditional generation independently. For video generation, our LandV adopts the pretrained DiT~\cite{dit} backbone of Wan2.1~\cite{wan}, operating on video latents encoded by its 3D VAE.
Within this backbone, we reuse Wan2.1's spatiotemporal self-attention and cross-attention for intra-view and conditional modeling, respectively.
Building on this, to facilitate cross-view interaction, we introduce inter-view self-attention that aggregates features from neighboring cameras at nearby latent frames, with parameters initialized from pretrained self-attention layers.
The attention updates are formulated as:
\begin{equation}
\label{eq:video-attention}
\begin{aligned}
H^v &\leftarrow H^v+\mathrm{SA}_{\mathrm{intra}}^v(H^v),\\
H^v &\leftarrow H^v+\mathrm{SA}_{\mathrm{inter}}^v(H^v,H_{\mathcal N}^v),\\
H^v &\leftarrow H^v+\mathrm{CA}_{g}^v(H^v,C_g^v),
\quad g\in\{\mathrm{text},\mathrm{obj},\}.
\end{aligned}
\end{equation}
where $\mathrm{SA}$ and $\mathrm{CA}$ denote self-attention and cross-attention, respectively.
$H^v$ denotes the video token features, and $H_{\mathcal N}^v$ represents neighboring cross-view features.

For LiDAR generation, we directly model two-channel range maps containing normalized range values and return validity, avoiding an intractable LiDAR VAE~\cite{sgldm,ultralidar}.
Unlike perspective images, LiDAR range maps are sparse and capture a 360-degree panoramic view, where the two ends of the azimuthal dimension are spatially adjacent. Given these characteristics, we employ spatio-temporal (ST) patch embedding with a patch size of $(2,2,8)$ and a lightweight DiT backbone with one-third the depth of the video branch, as their sparsity requires less compression and modeling capacity. To model the panoramic structure, we adopt temporal and azimuthal rotary positional embeddings (RoPE)~\cite{rope}, using integer harmonics to ensure $2\pi$-periodicity along the azimuthal dimension. Meanwhile, to efficiently handle the full azimuthal span, we partition self-attention into azimuthal windows and alternate regular and shifted windows across layers to enable cross-window interactions~\cite{swin}.
The attention updates are formulated as:
\begin{equation}
\label{eq:lidar-attention}
\begin{aligned}
H^l &\leftarrow H^l+\mathrm{SA}_{\mathrm{win}}^l(H^l),\\
H^l &\leftarrow H^l+\mathrm{CA}_{g}^l(H^l,C_g^l),
\quad g\in\{\mathrm{text},\mathrm{obj},\}.
\end{aligned}
\end{equation}
where $H^l$ denotes the LiDAR token features.

\noindent\textbf{Geometry-Guided Cross-Modal Interaction.}
To enable effective cross-modal interaction between the two streams, we present geometry-guided bidirectional cross-attention.
Specifically, sensor calibration establishes spatial correspondences between each camera view and the overlapping region in the LiDAR range map. At selected layers of both DiT blocks, bidirectional cross-modal attention is applied between corresponding regions, with each stream using its own features as queries and the other stream's features as keys and values.
Compared with global cross-modal attention, this blockwise design focuses feature exchange on geometrically relevant regions while reducing computational cost.
The bidirectional updates are formulated as:
\begin{equation}
\label{eq:cross-modal-attention}
\begin{aligned}
\widetilde H^v
&=H^v+\mathrm{CA}_{v\leftarrow l}
\left(H^v,H^l;\mathcal B_{v\leftarrow l}\right),\\
\widetilde H^l
&=H^l+\mathrm{CA}_{l\leftarrow v}
\left(H^l,H^v;\mathcal B_{l\leftarrow v}\right),
\end{aligned}
\end{equation}
where $\mathcal B_{v\leftarrow l}$ and $\mathcal B_{l\leftarrow v}$ denote the geometrically corresponding interaction blocks for the two directions.

\noindent\textbf{Training Objectives.}
Our LandV jointly trains both streams under a flow-matching framework~\cite{lipman2023flow}, where noisy inputs are constructed as $X_\tau^m=(1-\tau)X_0^m+\tau\epsilon^m$ with $m\in\{v,l\}$ and $\epsilon^m\sim\mathcal{N}(0,I)$. The video stream follows Wan2.1~\cite{wan} with the velocity-prediction objective. Since the LiDAR representation includes range values and return validity, we adopt x-prediction~\cite{li2026back} for range reconstruction and an auxiliary binary cross-entropy (BCE) loss for validity prediction. The LiDAR loss is formulated as:
\begin{equation}
\label{eq:lidar-generation-loss}
\mathcal L_{\mathrm{lidar}}
=\mathbb E\left[
\mathrm{MSE}_{M^l}(\hat D^l,D^l)
+\mathrm{BCE}(\hat M^l,M^l)
\right],
\end{equation}
where $\hat D^l$ and $\hat M^l$ denote the predicted normalized ranges and return-validity probabilities, respectively. $D^l$ and $M^l$ are their ground truths. The masked MSE is computed only over valid LiDAR returns. To further refine local geometry across patch boundaries in VAE-free LiDAR generation, we introduce the LGR head, as detailed in Sec.~\ref{sec:local-refinement}.

\subsection{Local Geometry Refinement (LGR) Head}
\label{sec:local-refinement}

To compensate for the absence of a dedicated spatial decoder in our VAE-free LiDAR branch, our LandV develops a Local Geometry Refinement (LGR) head to reconstruct local geometric structures from DiT patch tokens. As illustrated in Fig.~\ref{fig:landv-framework}(b), the LGR head leverages local spatial cues from noisy LiDAR inputs to complement backbone features, enabling fine-grained geometric refinement on the original range-map. 
Specifically, the final DiT tokens are linearly projected and unpatchified to obtain backbone features $F^l\in\mathbb{R}^{8\times T\times H\times W}$ and a base prediction $Y_{\mathrm{base}}^l\in\mathbb{R}^{2\times T\times H\times W}$. In parallel, the noisy LiDAR input $X_\tau^l$ is concatenated with a known-frame mask $K^l$ to form $\widetilde X_\tau^l=[X_\tau^l;K^l]$, from which two spatial convolutional layers extract local features $F_{\mathrm{loc}}^l\in\mathbb{R}^{16\times T\times H\times W}$. Here, $T$, $H$, and $W$ denote the temporal, beam, and azimuthal dimensions, respectively. These three feature maps are concatenated and fused to predict a residual correction:
\begin{equation}
\label{eq:local-refinement}
\begin{aligned}
F_{\mathrm{loc}}^l &= E_{\mathrm{loc}}^l(\widetilde X_\tau^l),\\
F_{\mathrm{mix}}^l &= E_{\mathrm{fuse}}^l([F_{\mathrm{loc}}^l;F^l;Y_{\mathrm{base}}^l]),\\
\hat Y^l &= Y_{\mathrm{base}}^l + R_{\mathrm{loc}}^l(F_{\mathrm{mix}}^l),
\end{aligned}
\end{equation}
where $E_{\mathrm{loc}}^l$ consists of two spatial convolutional layers with SiLU activations, $E_{\mathrm{fuse}}^l$ applies a depthwise convolution to integrate local spatial information, and $R_{\mathrm{loc}}^l$ employs pointwise convolutions with SiLU activations to fuse channel information and predict a two-channel residual.

\subsection{Adaptive Reciprocal Alignment (ARA) Loss}
\label{sec:attention-alignment}

\begin{table*}[!t]
\centering
\caption{Generation quality on nuScenes, grouped by generated modality. \textcolor{black!55}{\textbf{Gray entries}} denote results from the original papers that have not been verified under the same evaluation protocol. DriveDreamer-2 is evaluated with RGB conditioning. $\dagger$ indicates a different training dataset. MMD is multiplied by $10^4$. Global and Object measure scene- and object-level CGQ, respectively. -- denotes an inapplicable metric. Best and second-best results under the unified evaluation protocol are bold and underlined, respectively.}
\label{tab:generation-quality}
\small
\setlength{\tabcolsep}{6pt}
\renewcommand{\arraystretch}{1.0}
\begin{tabular}{@{}clcccccc@{}}
\toprule
\multirow{2}{*}{Modality} & \multirow{2}{*}{Method} & \multicolumn{2}{c}{Video quality} & \multicolumn{2}{c}{LiDAR quality} & \multicolumn{2}{c}{Cross-modal quality} \\
\cmidrule(lr){3-4}\cmidrule(lr){5-6}\cmidrule(l){7-8}
 & & FID $\downarrow$ & FVD $\downarrow$ & MMD $\downarrow$ & JSD $\downarrow$ & Global $\uparrow$ & Object $\uparrow$ \\
\midrule
\multirow{5}{*}{Video} & MagicDrive-V2~\cite{magicdrivev2} & 22.50 & 92.81 & -- & -- & -- & -- \\
 & DiST-T~\cite{dist4d} & 14.79 & 40.73 & -- & -- & -- & -- \\
 & DriveDreamer-2~\cite{drivedreamer2} & 11.91 & 71.22 & -- & -- & -- & -- \\
 & OmniNWM~\cite{omninwm} & 16.25 & 51.54 & -- & -- & -- & -- \\
 & \textbf{Ours (Video-only)} & \underline{9.81} & \underline{38.89} & -- & -- & -- & -- \\
\midrule
\multirow{5}{*}{LiDAR} & RangeLDM~\cite{rangeldm} & -- & -- & 0.94 & 0.084 & -- & -- \\
 & OpenDWM (DiT3)~\cite{opendwm} & -- & -- & 0.76 & 0.069 & -- & -- \\
 & OpenDWM (MaskGIT)~\cite{opendwm} & -- & -- & 1.36 & 0.119 & -- & -- \\
 & LaGen~\cite{lagen} & -- & -- & \underline{0.62} & \underline{0.068} & -- & -- \\
 & \textbf{Ours (LiDAR-only)} & -- & -- & 0.94 & 0.079 & -- & -- \\
\midrule
\multirow{7}{*}{Multimodal} & UniScene~\cite{uniscene} & 12.04 & 56.38 & 4.62 & 0.176 & 61.22 & 54.94 \\
 & UniScene (GT occ)~\cite{uniscene} & 13.14 & 74.52 & 3.43 & 0.145 & \underline{64.01} & \underline{61.26} \\
 & UniSceneV2~\cite{uniscenev2}$^{\dagger}$ & 24.84 & 111.74 & 2.32 & 0.132 & 28.90 & 31.28 \\
 & \muted{UniDriveDreamer}~\cite{unidrivedreamer} & \muted{2.81} & \muted{11.44} & \muted{0.27} & \muted{0.039} & \muted{--} & \muted{--} \\
 & \muted{Genesis}~\cite{genesis} & \muted{4.24} & \muted{16.95} & \muted{--} & \muted{--} & \muted{--} & \muted{--} \\
 & \muted{OmniGen}~\cite{omnigen} & \muted{21.01} & \muted{--} & \muted{2.94} & \muted{0.105} & \muted{--} & \muted{--} \\
 & \textbf{LandV (Ours)} & \textbf{7.57} & \textbf{30.45} & \textbf{0.30} & \textbf{0.058} & \textbf{75.68} & \textbf{70.16} \\
\bottomrule
\end{tabular}
\end{table*}

\begin{figure}[t]
    \centering
    \includegraphics[width=1.0\columnwidth]{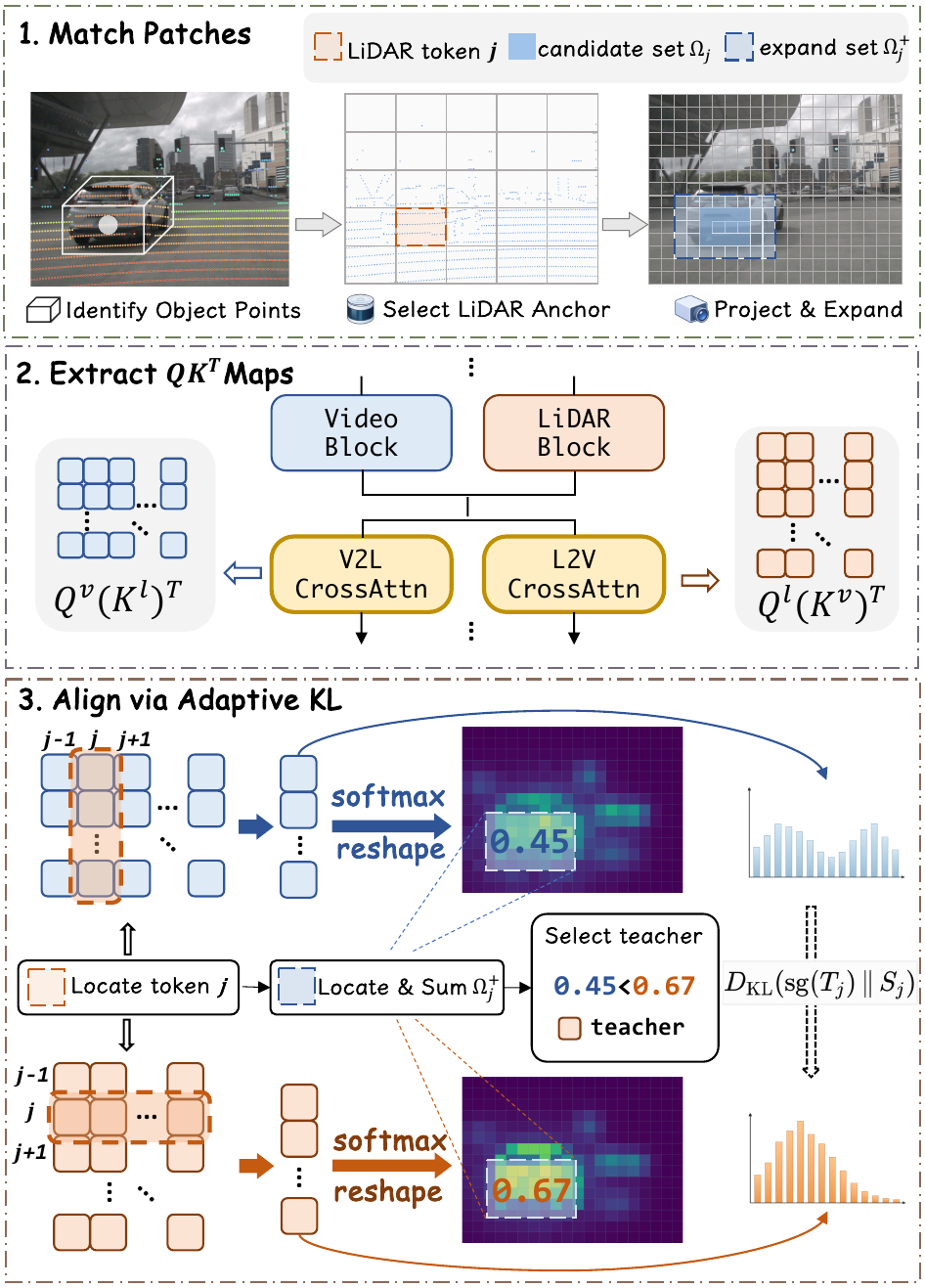}
    \caption{Overview of our ARA loss (see Sec. \ref{sec:attention-alignment} for details).}
    \label{fig:ara}
\end{figure}

While geometry-guided cross-attention restricts interactions to spatially corresponding regions, fine-grained object-level alignment remains to be learned. To enhance cross-modal object consistency, we introduce the Adaptive Reciprocal Alignment (ARA) loss, as illustrated in Fig.~\ref{fig:ara}.

During training, our ARA first identifies valid 3D object boxes and selects one central LiDAR token from each to form a set of geometric anchors $\mathcal A$. 
For each anchor $j\in\mathcal A$, we obtain its corresponding video region $\Omega_j$ by projecting LiDAR returns from the same object onto the video token grid.
The region is then expanded to $\Omega_j^+$ to provide a geometric reference for evaluating bidirectional cross-attention.
Next, using the selected anchor $j$ as an index, we extract the corresponding row and column from the LiDAR-to-video and video-to-LiDAR query--key similarity maps, respectively. The two vectors are normalized into distributions $p_j$ and $q_j$, each characterizing the LiDAR anchor's affinities with different video tokens. Then, the scores of both distributions are summed over $\Omega_j^+$:

\begin{equation}
\label{eq:alignment-mass}
m_j^p=\sum_{i\in\Omega_j^+}p_j(i),\qquad
m_j^q=\sum_{i\in\Omega_j^+}q_j(i).
\end{equation}
The distribution with the higher aggregated score is selected as the teacher $T_j$, while the other serves as the student $S_j$:

\begin{equation}
\label{eq:teacher-selection}
(T_j,S_j)=
\begin{cases}
(p_j,q_j), & m_j^p \geq m_j^q,\\
(q_j,p_j), & m_j^p < m_j^q.
\end{cases}
\end{equation}
We then align them through KL divergence, applying the stop-gradient operator $\operatorname{sg}(\cdot)$ to the teacher:
\begin{equation}
\label{eq:alignment-loss}
\mathcal L_{\mathrm{ARA}}
=\frac{1}{|\mathcal A|}
\sum_{j\in\mathcal A}
D_{\mathrm{KL}}
\left(\operatorname{sg}(T_j)\,\Vert\,S_j\right).
\end{equation}

Rather than enforcing hard token-level correspondences, ARA adaptively distills the soft attention distribution from the more geometrically aligned direction to its counterpart. The overall training objective combines the modality-specific generation losses and ARA loss:
\begin{equation}
\label{eq:joint-training-objective}
\mathcal L
=\mathcal L_{\mathrm{video}}
+\lambda_{\mathrm{lidar}}\mathcal L_{\mathrm{lidar}}
+\lambda_{\mathrm{ARA}}\mathcal L_{\mathrm{ARA}},
\end{equation}
where $\lambda_{\mathrm{lidar}}$ and $\lambda_{\mathrm{ARA}}$ are the corresponding loss weights.

\section{Experiments}
\label{sec:experiments}

\subsection{Experimental Setup}
\label{sec:experimental-setup}

\noindent\textbf{Dataset.}
We conduct experiments on nuScenes~\cite{nuscenes}, using 700 scenes for training and 150 for validation.
Following prior works~\cite{magicdrivev2,unidrivedreamer}, the 3D bounding-box annotations are interpolated into 12\,Hz to construct aligned 17-frame clips of six-view video and LiDAR.
Given scene conditions and first-frame observations subject to random dropout during training, our LandV generates the subsequent 16 frames.

\noindent\textbf{Benchmark.}
We introduce LandV-Bench, a benchmark for driving scene generation covering four dimensions: video quality, LiDAR quality, cross-modal geometric quality, and object appearance consistency.
The first two dimensions adopt established metrics: Fr\'echet Inception Distance (FID)~\cite{fid} and Fr\'echet Video Distance (FVD)~\cite{fvd} for video, and Maximum Mean Discrepancy (MMD) and Jensen--Shannon distance (JSD) for LiDAR~\cite{lidardm}.
For the latter two dimensions, we introduce Cross-modal Geometric Quality (CGQ) and Object Consistency Pass Rate (OCPR).
Specifically, CGQ evaluates the agreement between Pi3X-estimated video depths~\cite{pi3} and LiDAR-projected depths at both scene (Global) and object (Object) levels.
OCPR measures the proportion of object crop pairs judged consistent across views and time by a frozen Qwen3.5-9B~\cite{qwen35} model.
We benchmark state-of-the-art video-only~\cite{magicdrivev2,dist4d,drivedreamer2,omninwm}, LiDAR-only~\cite{rangeldm,opendwm,lagen}, and multimodal generation~\cite{uniscene,uniscenev2} methods using their publicly released implementations. For methods with shorter generation horizons, we extend their predictions to the benchmark's 16-frame horizon through autoregressive rollouts~\cite{uniscene,drivedreamer2}.
More details are provided in the supplementary material.

\begin{figure*}[t]
    \centering
    \includegraphics[width=\textwidth]{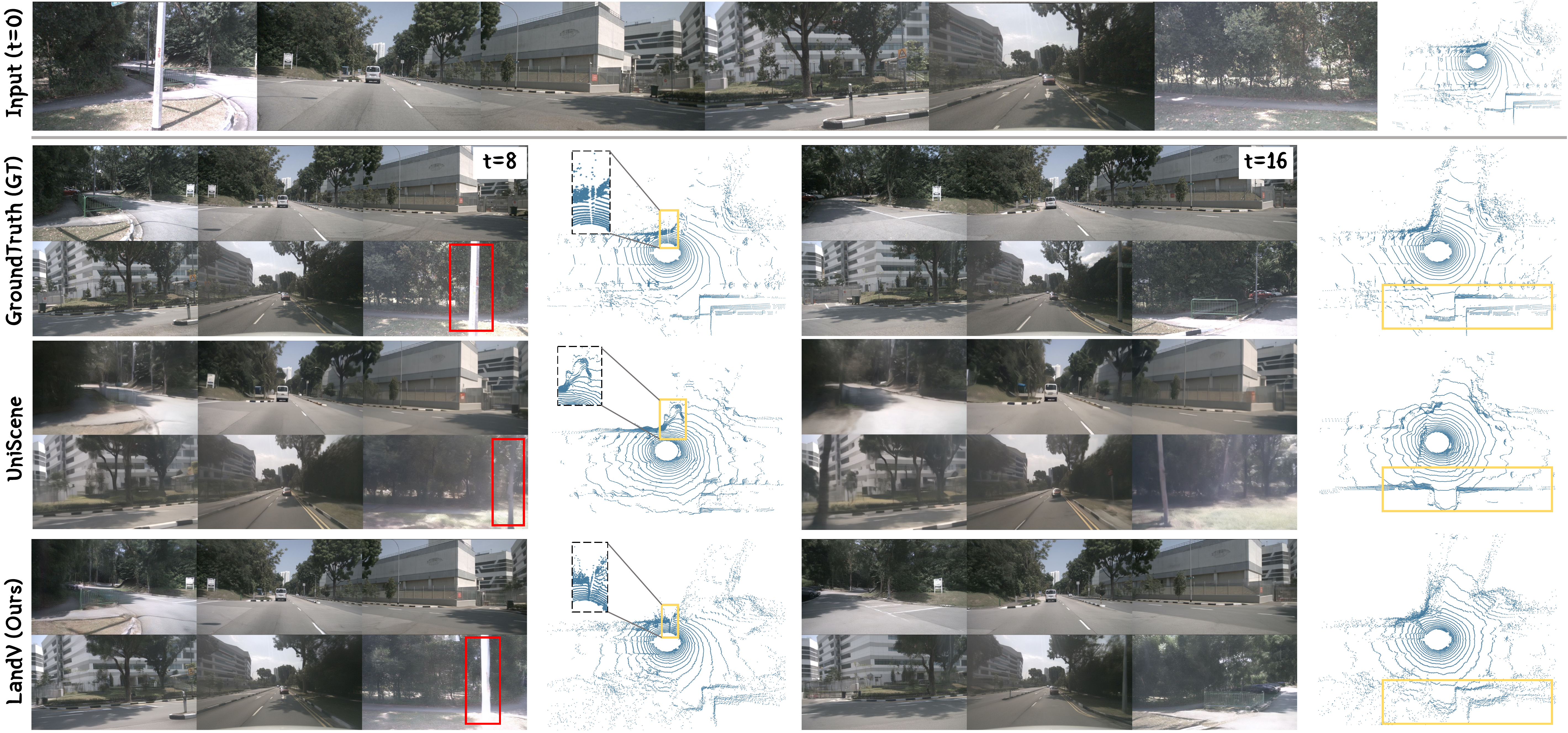}
    \caption{Qualitative comparison of LandV and UniScene~\cite{uniscene} on video and LiDAR generation.}
    \label{fig:qualitative-comparison}
\end{figure*}

\noindent\textbf{Implementation details.}
Our LandV processes videos at $480\times832$ resolution and LiDAR as $32\times1024$ range maps.
We train LandV for 10 epochs, switching from independent branch training to joint optimization in the second epoch.
Our single-modality baselines are also trained for 10 epochs to match the per-branch training budget.
More implementation details are provided in the supplementary material.

\subsection{Quantitative Results}
\label{sec:quantitative-results}

As shown in Tab.~\ref{tab:generation-quality}, we compare LandV against state-of-the-art video-only~\cite{magicdrivev2,dist4d,drivedreamer2,omninwm}, LiDAR-only~\cite{rangeldm,opendwm,lagen}, and multimodal generation methods~\cite{uniscene,uniscenev2,unidrivedreamer,genesis,omnigen} on LandV-Bench.
We also include a UniScene~\cite{uniscene} variant conditioned on ground-truth occupancy.

We evaluate video and LiDAR generation quality using FID/FVD and MMD/JSD, respectively.
Our LandV surpasses the evaluated single-modal baselines across all four metrics.
For video, it achieves an FID of 7.57 and an FVD of 30.45, improving by 36.4\% and 25.2\% over DriveDreamer-2~\cite{drivedreamer2} and DiST-T~\cite{dist4d}, the strongest video-only methods on the respective metrics.
For LiDAR, LandV outperforms LaGen~\cite{lagen}, reducing MMD from 0.62 to 0.30 and JSD from 0.068 to 0.058.
Moreover, compared with our separately trained video-only and LiDAR-only models, our LandV reduces FVD from 38.89 to 30.45 and MMD from 0.94 to 0.30, suggesting that joint modeling benefits both modalities.
Our LandV also achieves competitive generation quality among multimodal methods.
Among multimodal methods, LandV reduces FVD and MMD by 46.0\% and 93.5\%, respectively, relative to the occupancy-based UniScene~\cite{uniscene}.
For methods~\cite{unidrivedreamer,genesis,omnigen} without released implementations, we rely on their reported results, which have not been verified under the same evaluation protocol.
LandV achieves lower FID (7.57 vs. 21.01), MMD (0.30 vs. 2.94), and JSD (0.058 vs. 0.105) than those reported for the shared-backbone method OmniGen~\cite{omnigen}.
These results demonstrate that our LandV achieves strong generation quality for both video and LiDAR, with improvements over separately trained single-modal baselines. Its competitive performance against multimodal methods further suggests that modality-specific modeling and geometry-guided cross-modal interaction can be effectively combined within an asymmetric dual-stream architecture.

To further evaluate generated-scene consistency, we employ two metrics from LandV-Bench: CGQ measures geometric agreement between video and LiDAR at the scene and object levels, while OCPR assesses object appearance consistency across views and time.
As shown in Tab.~\ref{tab:generation-quality}, LandV achieves Global and Object CGQ scores of 75.68\% and 70.16\%, exceeding UniScene~\cite{uniscene}(with GT occupancy) by 11.67 and 8.90 percentage points, respectively.
In terms of object appearance consistency, LandV achieves an overall OCPR of 78.0\%, with leading cross-time and joint scores of 86.0\% and 82.0\%, respectively, as shown in Tab.~\ref{tab:ocpr}.
Compared with our video-only variant, joint generation improves overall OCPR from 50.0\% to 78.0\%, with gains of 36.0 and 46.0 percentage points in cross-view and joint consistency, respectively.
These results highlight LandV's improved cross-modal geometric consistency, supported by geometry-guided bidirectional cross-attention and ARA for effective cross-modal interaction and object-level alignment. Notably, while cross-time consistency remains comparable to the video-only variant, the substantial improvements in cross-view and joint consistency suggest that geometric cues from LiDAR help establish object correspondences across changing viewpoints, thereby preserving object appearance consistency.

\begin{table}[!t]
\centering
\caption{OCPR (\%, $\uparrow$) for cross-view, cross-time, and joint (different views at different times) consistency. Best and second-best results are shown in bold and underlined, respectively. DD2 denotes DriveDreamer-2~\cite{drivedreamer2}.}
\label{tab:ocpr}
\small
\setlength{\tabcolsep}{2pt}
\renewcommand{\arraystretch}{1.15}
\begin{tabular*}{\linewidth}{@{\extracolsep{\fill}}lcccc@{}}
\toprule
Method & Cross-view & Cross-time & Joint & Overall \\
\midrule
MagicDrive-V2~\cite{magicdrivev2} & 42.0 & 72.0 & 54.0 & 56.0 \\
DiST-T~\cite{dist4d} & 26.0 & 68.0 & 38.0 & 44.0 \\
DD2~\cite{drivedreamer2} & \textbf{70.0} & 82.0 & \underline{74.0} & \underline{75.3} \\
OmniNWM~\cite{omninwm} & 14.0 & 72.0 & 38.0 & 41.3 \\
UniScene~\cite{uniscene} & 18.0 & 82.0 & 50.0 & 50.0 \\
\midrule
\textbf{Ours (Video-only)} & 30.0 & \underline{84.0} & 36.0 & 50.0 \\
\textbf{LandV (Ours)} & \underline{66.0} & \textbf{86.0} & \textbf{82.0} & \textbf{78.0} \\
\bottomrule
\end{tabular*}
\end{table}

\subsection{Qualitative Results}
\label{sec:qualitative-results}

\begin{figure}[t]
    \centering
    \includegraphics[width=\linewidth]{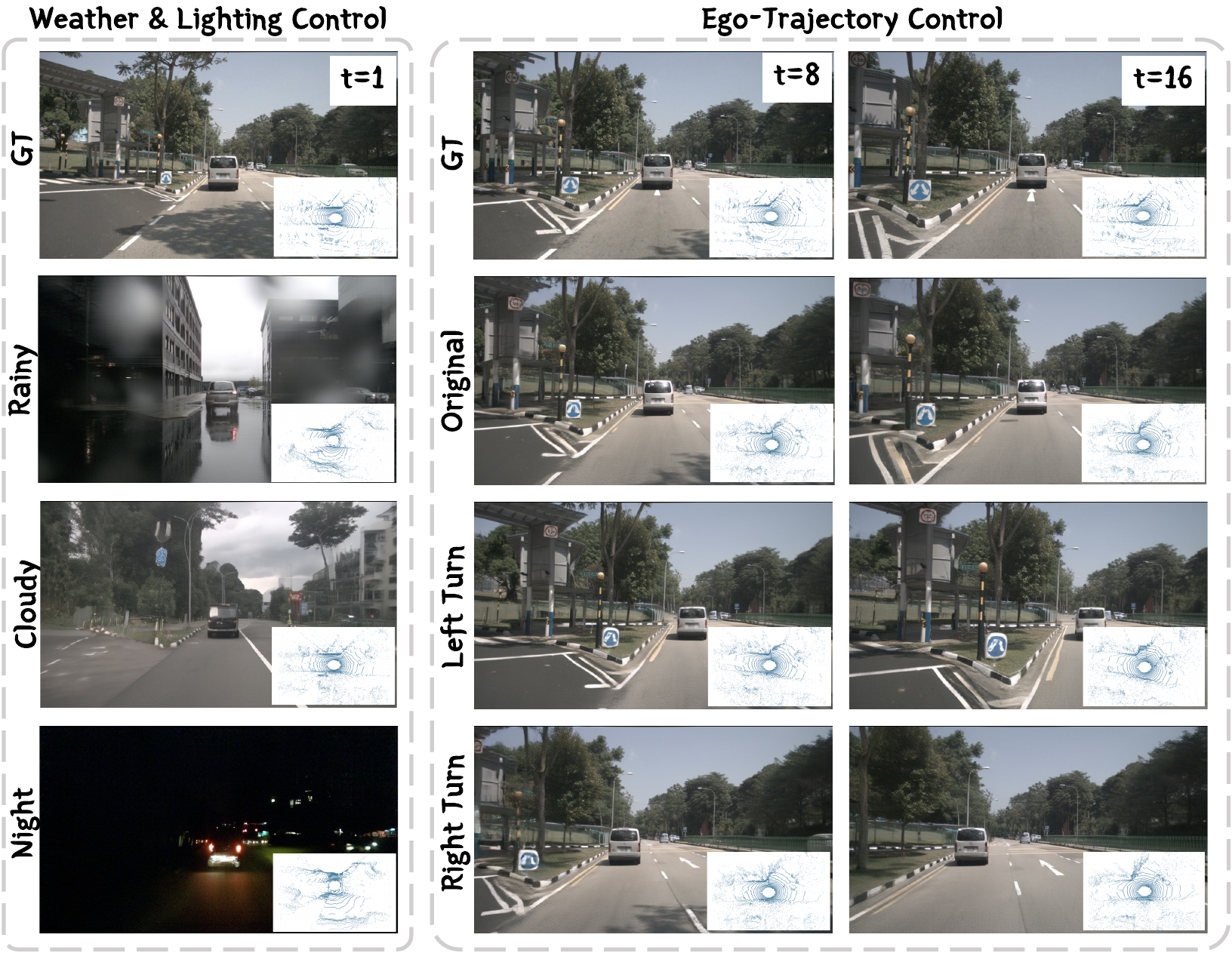}
    \caption{Controllable generation with LandV: weather and lighting variations (left) and ego-trajectory control (right).}
    \label{fig:controllable-generation}
\end{figure}

Figure~\ref{fig:qualitative-comparison} compares the video and LiDAR outputs of LandV and UniScene~\cite{uniscene} at $t=8$ and $t=16$. As highlighted in the red boxes, our LandV better preserves the appearance of the streetlight, demonstrating improved appearance consistency through effective cross-modal interaction.
Meanwhile, the yellow boxes show that LandV reconstructs more faithful LiDAR structures for both the streetlight and the building at $t=16$, suggesting that visual information also benefits LiDAR geometry generation.
Together, these results demonstrate the advantages of our asymmetric dual-stream architecture in leveraging complementary information from video and LiDAR to improve generation quality in both modalities.
Furthermore, we evaluate our LandV's controllable generation capabilities as a world model under two settings: weather and lighting variations, and novel-view generation through left- and right-turn ego trajectories.
As shown in Figure~\ref{fig:controllable-generation}, our LandV generates corresponding video and LiDAR observations under these controls, suggesting its potential to augment real-world datasets with corner cases that are difficult to collect.

\subsection{Downstream Evaluation}
\label{sec:downstream-evaluation}

We further evaluate LandV on downstream perception tasks following UniScene~\cite{uniscene}, augmenting the real training set with an equal number of generated video--LiDAR samples while matching training iterations and batch sizes.
As shown in Tab.~\ref{tab:downstream}, our LandV improves mAP from 65.56\% to 68.93\% and NDS from 69.76\% to 71.94\%, outperforming UniScene~\cite{uniscene} augmentation by 1.04 and 0.76 percentage points.
These results demonstrate the utility of LandV-generated multimodal data for downstream perception.
\begin{table}[!t]
\centering
\caption{3D detection results (\%) of BEVFusion~\cite{bevfusion} with various data augmentations on nuScenes~\cite{nuscenes}. Parentheses indicate gains over Real only (percentage points). Best results are in bold.}
\label{tab:downstream}
\small
\setlength{\tabcolsep}{2pt}
\renewcommand{\arraystretch}{1.15}
\begin{tabular*}{\linewidth}{@{\extracolsep{\fill}}lcc@{}}
\toprule
Training data & mAP $\uparrow$ & NDS $\uparrow$ \\
\midrule
Real only & 65.56 & 69.76 \\
Real + UniScene & 67.89 {\scriptsize\textcolor{green!40!black}{(+2.33)}} & 71.18 {\scriptsize\textcolor{green!40!black}{(+1.42)}} \\
Real + LandV (Ours) & \textbf{68.93 {\scriptsize\textcolor{green!40!black}{(+3.37)}}} & \textbf{71.94 {\scriptsize\textcolor{green!40!black}{(+2.18)}}} \\
\bottomrule
\end{tabular*}
\end{table}

\subsection{Ablation Studies}
\label{sec:ablation-studies}

\begin{table}[!t]
\centering
\caption{Ablation studies of modeling variants and core components (see Sec. \ref{sec:ablation-studies} for details). Best results are shown in bold.}
\label{tab:ablations}
\footnotesize
\setlength{\tabcolsep}{2pt}
\renewcommand{\arraystretch}{1.2}
\begin{tabular*}{\linewidth}{@{\extracolsep{\fill}}lcccccccc@{}}
\toprule
\multirow{2}{*}{Variant} & \multicolumn{2}{c}{Modules} & \multicolumn{2}{c}{Video $\downarrow$} & \multicolumn{2}{c}{LiDAR $\downarrow$} & \multicolumn{2}{c}{Geometry $\uparrow$} \\
\cmidrule(lr){2-3}\cmidrule(lr){4-5}\cmidrule(lr){6-7}\cmidrule(l){8-9}
& LGR & ARA & FID & FVD & MMD & JSD & Global & Object \\
\midrule

Unified & -- & -- & 12.07 & 51.03 & 1.29 & 0.160 & 64.33 & 59.04 \\
\midrule
\multirow{4}{*}{LandV} & -- & -- & 8.03 & 33.90 & 0.69 & 0.072 & 68.34 & 63.01 \\
& $\checkmark$ & -- & \textbf{7.51} & 32.20 & 0.38 & \textbf{0.057} & 70.83 & 64.07 \\
& -- & $\checkmark$ & 7.99 & 31.29 & 0.51 & 0.069 & 73.92 & 69.51 \\
\cmidrule(l){2-9}
& $\checkmark$ & $\checkmark$ & 7.57 & \textbf{30.45} & \textbf{0.30} & 0.058 & \textbf{75.68} & \textbf{70.16} \\
\bottomrule
\end{tabular*}
\end{table}

Finally, we ablate the modality-specific dual-stream design, LGR, and ARA under matched training budgets.
As shown in Tab.~\ref{tab:ablations}, the unified baseline, which uses VAE latents, a shared backbone, and v-prediction for both modalities, exhibits markedly inferior generation quality, particularly for LiDAR.
In contrast, our asymmetric dual-stream design substantially reduces MMD from 1.29 to 0.69 and JSD from 0.160 to 0.072, while also improving video generation quality, highlighting the critical importance of modality-specific modeling.
Furthermore, LGR improves LiDAR geometry, reducing MMD from 0.69 to 0.38, while ARA strengthens cross-modal alignment, improving Global and Object CGQ from 68.34\% and 63.01\% to 73.92\% and 69.51\%, respectively.
Combining both components achieves the lowest MMD of 0.30 and the highest Global and Object CGQ scores of 75.68\% and 70.16\%.
\section{Conclusion}
\label{sec:conclusion}

In this paper, we proposed LandV, an asymmetric dual-stream multimodal world model for autonomous driving.
For modality-specific modeling, the architecture pairs a pretrained latent video stream with a lightweight, VAE-free LiDAR stream, using distinct prediction objectives and a Local Geometry Refinement (LGR) head to capture fine-grained LiDAR geometry.
To enable effective cross-modal interaction, our LandV combines geometry-guided bidirectional cross-attention with an Adaptive Reciprocal Alignment (ARA) loss for object-level alignment.
Experiments on LandV-Bench demonstrate competitive performance in video and LiDAR generation, cross-modal geometric quality, and object appearance consistency across views and time.
These results support the effectiveness of combining modality-specific modeling with direct cross-modal interaction in a dual-stream architecture and suggest its potential for supporting downstream autonomous driving tasks.

\section*{LLM Usage Statement}
We used OpenAI GPT models, including the Codex assistant, to assist with code optimization and hyperparameter tuning, and to refine the language, organization, and presentation of the manuscript. This assistance also covered LaTeX typesetting and figure layout. A frozen Qwen3.5-9B model was used to assess object appearance consistency for OCPR, as described in Sec.~\ref{sec:experimental-setup}. The authors conceived the research motivation, designed the algorithm, developed the mathematical derivations, and designed and conducted the experiments. The authors directed the use of these tools and take full responsibility for the scientific claims, results, and final manuscript.

{
    \small
    \bibliographystyle{ieeenat_fullname}
    \bibliography{main}
}

\end{document}

\appendix
\renewcommand{\thepage}{S\arabic{page}}
\setcounter{page}{1}
\setcounter{table}{0}
\renewcommand{\thetable}{S\arabic{table}}
\renewcommand{\theHtable}{supp.\arabic{table}}
\setcounter{equation}{0}
\renewcommand{\theequation}{S\arabic{equation}}
\renewcommand{\theHequation}{supp.\arabic{equation}}
\renewcommand{\thefigure}{S\arabic{figure}}
\maketitlesupplementary
\setlength{\dbltextfloatsep}{8pt plus 2pt minus 2pt}
\setlength{\textfloatsep}{9pt plus 2pt minus 2pt}
\setlength{\dblfloatsep}{7pt plus 1pt minus 1pt}
\noindent
In the supplementary material, we provide the following additional details on our method and experiments:
\begin{itemize}
	\item \textbf{Implementation Details} (Section~\ref{sec:supp-implementation}): LandV training data, model initialization, joint optimization, and inference settings for video and LiDAR generation.
	
	\item \textbf{Benchmark Details} (Section~\ref{sec:supp-benchmark}): Evaluation protocols for video and LiDAR generation, cross-modal geometric agreement, and object consistency across views and time.
	
	\item \textbf{Additional Quantitative Experiments} (Section~\ref{sec:supp-quantitative}): Ablations on the LandV architecture, local geometry refinement, loss weights, and adaptive reciprocal alignment.
	
	\item \textbf{Additional Qualitative Experiments} (Section~\ref{sec:supp-qualitative}): Comparisons with UniScene~\cite{uniscene} and single-modality models, with examples of controllable generation.
	
	\item \textbf{OCPR Prompts} (Section~\ref{sec:supp-ocpr-prompts}): Prompts for object recognition and appearance consistency across views and time.
\end{itemize}

\section{Implementation Details}
\label{sec:supp-implementation}

\noindent\textbf{Training data and initialization.}
We train LandV on 153,473 clips from all 700 nuScenes~\cite{nuscenes} training scenes.
Each clip has 17 synchronized frames at 12\,Hz~\cite{magicdrivev2,unidrivedreamer}, with six $480\times832$ camera views and $32\times1024$ LiDAR range maps.
We initialize the video backbone from Wan2.1-T2V-1.3B~\cite{wan} and train the LiDAR backbone from scratch.
The video and LiDAR branches are trained independently for the first epoch, followed by nine epochs of joint training.

\smallskip
\noindent\textbf{Joint optimization.}
Training runs on 32 NVIDIA H100 GPUs with BF16 autocast and a global batch size of 64 (two clips per GPU).
We use AdamW~\cite{kingma2015adam,adamw} with constant learning rates and a weight decay of 0.01.
During training, first-frame observations are randomly dropped simultaneously for both modalities, and object appearance features are also randomly dropped.
When retained, the projected DINOv2~\cite{dinov2} feature is added to the learnable appearance embedding; otherwise, the embedding alone is used.
For joint training, we set $\lambda_{\mathrm{lidar}}=1$ and $\lambda_{\mathrm{ARA}}=0.01$.

\smallskip
\noindent\textbf{Inference.}
Given the initial six camera views and LiDAR sweep, our LandV jointly generates the next 16 frames of both modalities in one call.
We use 30 Euler steps with a shared noise schedule, a schedule shift of 5, and a guidance scale of 1.
For object appearance, we randomly select an out-of-window reference crop for each object when available and add its projected DINOv2~\cite{dinov2} feature to the learned appearance embedding.

\section{Benchmark Details}
\label{sec:supp-benchmark}

This section provides further details on the evaluation protocols of LandV-Bench, along with metric-specific supplementary experiments and validation studies.
For FID and FVD, we additionally report per-camera and matched-horizon comparisons (Section~\ref{sec:supp-fid-fvd}).
For MMD and JSD, we detail their computation for LiDAR generation (Section~\ref{sec:supp-mmd-jsd}).
For CGQ, we validate the RGB depth estimates against ground-truth LiDAR (Section~\ref{sec:supp-cgq}).
For OCPR, we validate the evaluator on ground-truth image pairs and report stage-wise pass rates (Section~\ref{sec:supp-ocpr}).

\subsection{FID and FVD}
\label{sec:supp-fid-fvd}
Following the FVD protocol of MagicDrive-V2~\cite{magicdrivev2}, we pool features across all six cameras for both FID~\cite{fid} and FVD~\cite{fvd}.
For the 17-frame evaluation of MagicDrive-V2~\cite{magicdrivev2}, DiST-T~\cite{dist4d}, OmniNWM~\cite{omninwm}, UniSceneV2~\cite{uniscenev2}, our LandV, and its video-only variant, we generate four six-view clips per validation window using different seeds. We exclude the initial conditioning frame and score the 16 future frames against ground truth. DriveDreamer-2~\cite{drivedreamer2} and UniScene~\cite{uniscene} each predict seven future frames per call, so the 16-frame evaluation in the main paper requires multiple calls. For subsequent calls, DriveDreamer-2~\cite{drivedreamer2} uses ground-truth RGB frames, whereas UniScene~\cite{uniscene} uses its own predictions. 
To compare methods with different output resolutions, we crop the generated and ground-truth frames to a common field of view and resize them to $224\times400$ pixels before scoring. 
Since MagicDrive-V2~\cite{magicdrivev2} pools features across all six camera views for FVD, we also report per-view FID and FVD. Table~\ref{tab:supp-video-views} reports the per-view scores, their mean, and the pooled scores. LandV ranks first on both metrics in all six views among these methods.

\begin{table*}[t]
	\centering
	\caption{Per-camera FID~\cite{fid} and FVD~\cite{fvd} on nuScenes~\cite{nuscenes}. Mean averages the six camera scores. Pooled combines features from all cameras before scoring. GT occ denotes ground-truth occupancy input. UniSceneV2~\cite{uniscenev2} was trained without nuScenes~\cite{nuscenes} data.}
	\label{tab:supp-video-views}
	\normalsize
	\renewcommand{\arraystretch}{1.1}
	\setlength{\tabcolsep}{5pt}
	\begin{tabular}{@{}llrrrrrrrr@{}}
		\toprule
		Metric & Method & Front L. & Front & Front R. & Rear R. & Rear & Rear L. & Mean & Pooled \\
		\midrule
		\multirow{9}{*}{FID $\downarrow$}
		& MagicDrive-V2~\cite{magicdrivev2}     & 48.30 & 43.11 & 52.12 & 55.84 & 38.89 & 53.23 & 48.58 & 22.50 \\
		& DiST-T~\cite{dist4d}                 & 30.82 & 22.62 & 28.67 & 31.28 & 24.85 & 33.29 & 28.59 & 14.79 \\
		& DriveDreamer-2~\cite{drivedreamer2}   & 27.27 & 20.44 & 25.98 & 27.72 & 21.08 & 28.94 & 25.24 & 11.91 \\
		& OmniNWM~\cite{omninwm}               & 34.45 & 23.08 & 34.19 & 39.54 & 21.21 & 37.29 & 31.63 & 16.25 \\
		& UniScene~\cite{uniscene}              & 30.00 & 22.81 & 29.17 & 30.96 & 22.62 & 32.91 & 28.08 & 12.04 \\
		& UniScene (GT occ)~\cite{uniscene}     & 33.41 & 24.94 & 32.59 & 34.86 & 24.82 & 37.57 & 31.37 & 13.14 \\
		& UniSceneV2~\cite{uniscenev2}          & 42.31 & 33.00 & 39.28 & 46.61 & 46.32 & 47.42 & 42.49 & 24.84 \\
		& Ours (Video-only)                    & 19.44 & 16.32 & 19.54 & 22.51 & 17.64 & 24.58 & 20.00 & 9.81 \\
		& LandV             & \textbf{17.86} & \textbf{12.84} & \textbf{17.63} & \textbf{20.30} & \textbf{13.59} & \textbf{21.53} & \textbf{17.29} & \textbf{7.57} \\
		\midrule
		\multirow{9}{*}{FVD $\downarrow$}
		& MagicDrive-V2~\cite{magicdrivev2}     & 261.78 & 299.86 & 281.63 & 215.69 & 343.60 & 229.74 & 272.05 & 92.81 \\
		& DiST-T~\cite{dist4d}                 & 116.58 & 164.65 & 129.81 & 118.05 & 187.38 & 116.36 & 138.81 & 40.73 \\
		& DriveDreamer-2~\cite{drivedreamer2}   & 143.77 & 209.39 & 154.35 & 141.26 & 236.10 & 132.70 & 169.59 & 71.22 \\
		& OmniNWM~\cite{omninwm}               & 136.86 & 154.68 & 145.29 & 135.79 & 168.78 & 132.11 & 145.58 & 51.54 \\
		& UniScene~\cite{uniscene}              & 121.63 & 174.05 & 126.25 & 122.14 & 198.86 & 114.36 & 142.88 & 56.38 \\
		& UniScene (GT occ)~\cite{uniscene}     & 144.30 & 198.32 & 158.28 & 157.34 & 222.01 & 147.24 & 171.25 & 74.52 \\
		& UniSceneV2~\cite{uniscenev2}          & 178.85 & 283.93 & 179.39 & 175.50 & 417.24 & 185.58 & 236.75 & 111.74 \\
		& Ours (Video-only)                    & 98.32 & 118.55 & 114.15 & 106.01 & 149.58 & 103.54 & 115.03 & 38.89 \\
		& LandV             & \textbf{79.93} & \textbf{98.62} & \textbf{91.55} & \textbf{86.12} & \textbf{136.34} & \textbf{80.75} & \textbf{95.55} & \textbf{30.45} \\
		\bottomrule
	\end{tabular}
\end{table*}

For baselines~\cite{drivedreamer2,uniscene}  that generate fewer frames per call, we additionally conduct comparisons at their native prediction horizons without multi-step rollout.
Specifically, we select LandV predictions and ground-truth frames matching the seven future timestamps of each baseline's first generation call.
FID is computed on the matched frames, while all clips are linearly resampled to nine frames for FVD to accommodate temporal pooling in I3D~\cite{i3d}.
As shown in Table~\ref{tab:supp-video-short}, LandV achieves lower FID and FVD than both DriveDreamer-2~\cite{drivedreamer2} and UniScene~\cite{uniscene}, demonstrating its consistent advantages across different prediction horizons.

\begin{table}[t]
	\centering
	\small
	\caption{LandV and each baseline at the seven future timestamps of each baseline's first call, across 150 windows and six views.}
	\label{tab:supp-video-short}
	\normalsize
	\renewcommand{\arraystretch}{1.16}
	\setlength{\tabcolsep}{1.8pt}
	\begin{tabular*}{\columnwidth}{@{\extracolsep{\fill}}llrr@{}}
		\toprule
		Future frames & Method & FID $\downarrow$ & FVD $\downarrow$ \\
		\midrule
		\multirow{2}{*}{\shortstack[l]{$t_3,t_5,t_7,t_{10},$\\$t_{12},t_{14},t_{16}$}} & DriveDreamer-2~\cite{drivedreamer2} & 18.62 & 76.72 \\
		& LandV & \textbf{10.32} & \textbf{29.36} \\
		\midrule
		\multirow{2}{*}{$t_1$--$t_7$} & UniScene~\cite{uniscene} & 18.04 & 78.88 \\
		& LandV & \textbf{9.07} & \textbf{22.13} \\
		\bottomrule
	\end{tabular*}
\end{table}

\subsection{MMD and JSD}
\label{sec:supp-mmd-jsd}
We follow LidarDM~\cite{lidardm} to compute MMD and JSD from bird's-eye-view (BEV) occupancy histograms.
Specifically, we voxelize each sweep in the LiDAR coordinate frame over $[-50,50]\times[-50,50]\times[-3.23,3.77]$\,m with voxel sizes of $(0.15625,0.15625,0.2)$\,m and apply adaptive average pooling to obtain a $100\times100$ BEV histogram from the binary occupancy volume.
For MMD~\cite{gretton2012mmd}, we normalize each sweep histogram and report the biased estimate of squared MMD using a Gaussian kernel ($\sigma=0.5$), scaled by $10^4$.
For JSD, we sum the unnormalized histograms within each set and compute the Jensen--Shannon distance~\cite{endres2003metric} between the two normalized sums.
We use these metrics to compare our LandV and its LiDAR-only variant with RangeLDM~\cite{rangeldm}, OpenDWM~\cite{opendwm} (DiT3 and MaskGIT), LaGen~\cite{lagen}, UniScene~\cite{uniscene} (generated or ground-truth occupancy), and UniSceneV2~\cite{uniscenev2}. For methods evaluated on 16 future sweeps, four generations per window across 150 windows yield 9,600 predicted sweeps per method for comparison with the same 2,400 ground-truth sweeps.

\begin{table}[t]
	\centering
	\small
	\caption{CGQ (\%) at $\delta_{1.25}$ on GT scenes (Global) and objects (Object). Aligned pairs synchronized RGB and LiDAR using all valid returns. Shifted changes the RGB timestamp within each view and uses returns valid in both pairings. Parentheses give changes in percentage points.}
	\label{tab:supp-cgq-gt}
	\normalsize
	\renewcommand{\arraystretch}{1.16}
	\setlength{\tabcolsep}{1.8pt}
	\begin{tabular*}{\columnwidth}{@{\extracolsep{\fill}}llrr@{}}
		\toprule
		Depth model & Scope & Aligned & Shifted \\
		\midrule
		\multirow{2}{*}{Pi3X~\cite{pi3}} & Global & \textbf{80.36} & 66.91~\textcolor{red!70!black}{($-13.45$)} \\
		& Object & \textbf{72.24} & 50.72~\textcolor{red!70!black}{($-21.52$)} \\
		\midrule
		\multirow{2}{*}{DA3-BASE~\cite{da3}} & Global & 57.84 & 49.69~\textcolor{red!70!black}{($-8.15$)} \\
		& Object & 47.67 & 36.07~\textcolor{red!70!black}{($-11.60$)} \\
		\bottomrule
	\end{tabular*}
\end{table}

\begin{figure*}[!t]
	\centering
	\includegraphics[width=\textwidth]{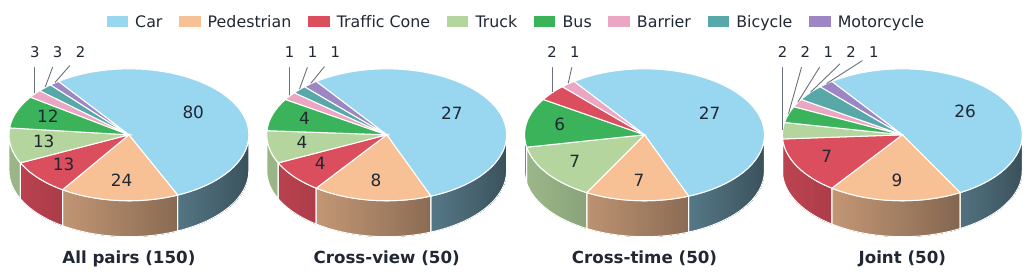}
	\caption{OCPR pair counts by object category, overall and by pairing type. Joint denotes cross-view \& time pairs.}
	\label{fig:supp-ocpr-classes}
\end{figure*}

\begin{table}[t]
	\centering
	\small
	\caption{OCPR on generated pairs (passed/evaluated). Stage 1 requires class recognition in both crops. Stage 2 requires consistent appearance in both orders. Bold marks the highest pass count. Overall OCPR is Stage 2 passes divided by 150. The last row omits object appearance references at inference.}
	\label{tab:supp-ocpr-stages}
	\normalsize
	\renewcommand{\arraystretch}{1.16}
	\setlength{\tabcolsep}{1.8pt}
	\begin{tabular*}{\columnwidth}{@{\extracolsep{\fill}}lcc@{}}
		\toprule
		\multirow{2}{*}{Method} & Stage 1 & Stage 2 \\
		& {\footnotesize All pairs} & {\footnotesize Stage 1 passes} \\
		\midrule
		MagicDrive-V2~\cite{magicdrivev2} & 86/150 & 84/86 \\
		DiST-T~\cite{dist4d} & 67/150 & 66/67 \\
		DriveDreamer-2~\cite{drivedreamer2} & \textbf{123}/150 & 113/123 \\
		OmniNWM~\cite{omninwm} & 64/150 & 62/64 \\
		UniScene~\cite{uniscene} & 81/150 & 75/81 \\
		\addlinespace[2pt]
		Ours (Video-only) & 112/150 & 75/112 \\
		LandV (Ours) & 119/150 & 117/119 \\
		LandV w/o object refs. & 120/150 & \textbf{118}/120 \\
		\bottomrule
	\end{tabular*}
\end{table}

\subsection{Cross-modal Geometric Quality (CGQ)}
\label{sec:supp-cgq}
We use Pi3X~\cite{pi3} to estimate depth from RGB images using camera intrinsics and poses, then recover metric depth by aligning the estimated camera centers with their calibrated counterparts. For each camera and time step, we project the corresponding LiDAR sweep into the image using the sensor calibration. At each valid projected pixel, we compare the video depth $d_v$ with the camera-space depth $d_l$ of the nearest LiDAR return. The depths agree at $\delta_{1.25}$ when $\max(d_v/d_l,d_l/d_v)<1.25$, meaning they differ by less than 25\% of the smaller depth. Global CGQ computes one agreement rate per image and averages these rates within each scene, then across scenes. Object CGQ measures agreement inside the annotated 3D boxes and averages over object frames with LiDAR support.

To test whether Pi3X~\cite{pi3} provides a reliable depth reference for evaluating RGB--LiDAR geometric agreement, we compare its depth estimates from real RGB images with synchronized LiDAR measurements projected into the same views across 150 nuScenes~\cite{nuscenes} validation scenes. As shown in Table~\ref{tab:supp-cgq-gt}, Pi3X~\cite{pi3} achieves higher agreement than DA3-BASE~\cite{da3} on both Global and Object CGQ, supporting its use as the depth estimator for CGQ.
We pair LiDAR sweeps and boxes at $(t_2,t_7,t_{15})$ with video depths from $(t_7,t_{15},t_2)$ in the same scene and camera. Both estimators yield lower Global and Object CGQ for these shifted pairs than for the original, temporally aligned pairs.

\subsection{Object Consistency Pass Rate (OCPR)}
\label{sec:supp-ocpr}

We construct an OCPR evaluation set of 150 object pairs from 86 nuScenes~\cite{nuscenes} validation scenes, with 50 pairs for each of three tasks. Cross-view pairs show the same object at the same time from different cameras. Cross-time pairs show the same object in the same camera four frames apart, while joint pairs show the same object in different cameras nine or eleven frames apart. Figure~\ref{fig:supp-ocpr-classes} shows the object category distributions overall and by task. Furthermore, we divide the pairs into small, medium, and large groups based on the smaller short side of their two crops, measured at $480\times832$ resolution before resizing. The group counts are 16/17/17 for cross-view, 17/16/17 for cross-time, and 17/17/16 for joint. The corresponding size thresholds are 75.09 and 110.16 pixels for cross-view, 73.42 and 103.24 pixels for cross-time, and 73.69 and 103.36 pixels for joint.

\begin{table}[t]
	\centering
	\small
	\caption{OCPR (\%) on GT crops. Each judge evaluates 150 same-instance pairs. For the negative control, crops from different instances of the same class form 150 pairs.}
	\label{tab:supp-ocpr-gt}
	\fontsize{9.3}{11}\selectfont
	\renewcommand{\arraystretch}{1.16}
	\setlength{\tabcolsep}{1.8pt}
	\begin{tabular*}{\columnwidth}{@{\extracolsep{\fill}}lrr@{}}
		\toprule
		Judge & Same instance & Different instances \\
		\midrule
		Qwen3.5-9B~\cite{qwen35} & \textbf{97.33} & 5.33 \\
		Qwen3-VL-8B-Instruct~\cite{qwen3vl} & 96.00 & \textemdash \\
		Step3-VL-10B~\cite{step3vl} & 94.00 & \textemdash \\
		\bottomrule
	\end{tabular*}
\end{table}

\begin{table*}[!t]
	\centering
	\caption{Allocation of 40 transformer blocks between the video and LiDAR branches (V/L). Parameter counts differ by 0.0355\%. Global and Object report scene- and object-level CGQ (\%). MMD is multiplied by $10^4$.}
	\label{tab:supp-branch-depth}
	\setlength{\tabcolsep}{6pt}
	\normalsize
	\renewcommand{\arraystretch}{1.16}
	\begin{tabular}{@{}lccccccc@{}}
		\toprule
		\multirow{2}{*}{Depth (V/L)} & \multirow{2}{*}{Params. (M)} & \multicolumn{2}{c}{Video quality} & \multicolumn{2}{c}{LiDAR quality} & \multicolumn{2}{c}{Cross-modal quality} \\
		\cmidrule(lr){3-4}\cmidrule(lr){5-6}\cmidrule(l){7-8}
		& & FID $\downarrow$ & FVD $\downarrow$ & MMD $\downarrow$ & JSD $\downarrow$ & Global $\uparrow$ & Object $\uparrow$ \\
		\midrule
		30/10 & 2,269.87 & \textbf{7.57} & \textbf{30.45} & \textbf{0.30} & \textbf{0.058} & \textbf{75.68} & \textbf{70.16} \\
		20/20 & 2,270.67 & 16.60 & 59.01 & 0.78 & 0.067 & 57.18 & 49.66 \\
		\bottomrule
	\end{tabular}
\end{table*}

\begin{table*}[!t]
\begin{minipage}[t]{\columnwidth}
    \vspace{0pt}
    \captionsetup{type=table}
	\centering
	\small
	\setlength{\tabcolsep}{3pt}
	\caption{LiDAR-only self-attention over alternating regular and shifted azimuth windows or the full panorama (Global). We report attention FLOPs and runtime for a single forward pass through one LiDAR self-attention layer. MMD is multiplied by $10^4$.}
	\label{tab:supp-lidar-self-attention}
	\normalsize
	\renewcommand{\arraystretch}{1.16}
	\setlength{\tabcolsep}{1.8pt}
	\begin{tabular*}{\columnwidth}{@{\extracolsep{\fill}}lrrcc@{}}
		\toprule
		Attention & MMD $\downarrow$ & JSD $\downarrow$ & FLOPs (T) $\downarrow$ & Time (ms) $\downarrow$ \\
		\midrule
		Shifted window & \textbf{0.94} & \textbf{0.079} & \textbf{0.522} & \textbf{8.10} \\
		Global & 1.38 & 0.092 & 2.087 & 14.20 \\
		\bottomrule
	\end{tabular*}
\end{minipage}\hfill
\begin{minipage}[t]{\columnwidth}
    \vspace{0pt}
    \captionsetup{type=table}
	\centering
	\small
	\setlength{\tabcolsep}{2.7pt}
	\caption{Cross-attention in LandV. Each camera view attends to its calibrated LiDAR region (Region-to-region) or the full panorama (Global spatial). Global and Object report scene-level and object-level CGQ (\%). FLOPs and time cover one bidirectional layer per forward pass.}
	\label{tab:supp-cross-modal-attention}
	\fontsize{9.2}{11}\selectfont
	\renewcommand{\arraystretch}{1.16}
	\setlength{\tabcolsep}{1.8pt}
	\begin{tabular*}{\columnwidth}{@{\extracolsep{\fill}}lrrcc@{}}
		\toprule
		Attention & Global $\uparrow$ & Object $\uparrow$ & FLOPs (T) $\downarrow$ & Time (ms) $\downarrow$ \\
		\midrule
		Region-to-region & \textbf{75.68} & \textbf{70.16} & \textbf{1.126} & \textbf{49.61} \\
		Global spatial & 72.63 & 67.29 & 5.653 & 63.24 \\
		\bottomrule
	\end{tabular*}
\end{minipage}
\end{table*}

For each pair, we project the ground-truth 3D boxes into the camera images to extract the corresponding object crops and evaluate them in two stages using a frozen Qwen3.5-9B model~\cite{qwen35}. In Stage~1, the evaluator checks each crop independently for a recognizable object of the expected category. If both pass, the evaluator assesses appearance consistency in Stage~2 twice, once for each ordering of the two crops. A pair passes only if the evaluator finds its appearance consistent across the two crops in both runs. Sections~\ref{sec:supp-ocpr-prompt-recognition} and~\ref{sec:supp-ocpr-prompt-consistency} list the Stage~1 and Stage~2 prompts, respectively. OCPR is computed as the percentage of all 150 pairs passing both stages. Table~\ref{tab:supp-ocpr-stages} reports the pass rate among pairs evaluated at each stage, with exact counts. Without object appearance references at inference, LandV passes 118 of the 150 OCPR pairs, showing that it can maintain consistent object appearance across generated views and frames without these references.

To assess evaluator reliability on the selected samples, we compare three vision-language models~\cite{qwen35,qwen3vl,step3vl} using the same two-stage protocol on ground-truth RGB crops. Since each pair contains the same annotated object instance, these ground-truth pairs serve as positive references for evaluating appearance consistency. As shown in Table~\ref{tab:supp-ocpr-gt}, Qwen3.5-9B~\cite{qwen35} achieves the highest pass rate of 97.33\% (146/150), compared with 96.00\% (144/150) for Qwen3-VL-8B-Instruct~\cite{qwen3vl} and 94.00\% (141/150) for Step3-VL-10B~\cite{step3vl}. To test whether the evaluator can also reject negative pairs, we re-pair GT crops from different instances of the same object class. With Qwen3.5-9B~\cite{qwen35}, OCPR drops by 92.00 percentage points to 5.33\% (8/150): only eight negative pairs pass the protocol.

\section{Additional Quantitative Experiments}
\label{sec:supp-quantitative}

We begin with the LandV architecture, examining branch depth, LiDAR self-attention, and the spatial scope of cross-modal attention (Section~\ref{sec:supp-arch}). We then focus on Local Geometry Refinement (LGR), studying how the design of the refinement head affects LiDAR generation (Section~\ref{sec:supp-lgr}). Next, we test sensitivity to the LiDAR and alignment loss weights (Section~\ref{sec:supp-loss-weights}). Finally, we examine teacher selection in Adaptive Reciprocal Alignment (ARA) and introduce ARA-fast, a lightweight variant that uses fewer tokens during training (Section~\ref{sec:supp-ara}).

\subsection{LandV architecture}
\label{sec:supp-arch}
\noindent\textbf{Branch depth.}
To examine how backbone depth should be distributed across modalities, we compare two allocations of 40 transformer blocks. The asymmetric model uses 30 video blocks and 10 LiDAR blocks, while the balanced model uses 20 blocks in each stream. Both models have approximately 2.27 billion parameters. As shown in Table~\ref{tab:supp-branch-depth}, the 30/10 model more closely matches the ground-truth distributions in both modalities. Relative to the 20/20 model, it reduces video FID from 16.60 to 7.57 and FVD from 59.01 to 30.45, while reducing LiDAR MMD from 0.78 to 0.30 and JSD from 0.067 to 0.058.

\smallskip
\begin{table*}[!t]
\begin{minipage}[t]{\columnwidth}
    \vspace{0pt}
    \captionsetup{type=table}
	\centering
	\small
	\caption{LiDAR-only refinement heads. The baseline uses one spatial convolution. Both LGR heads fuse local and backbone features, using two local convolutional layers by default or three narrower layers in the deeper extractor. MMD is multiplied by $10^4$.}
	\label{tab:supp-lgr-head}
	\normalsize
	\renewcommand{\arraystretch}{1.35}
	\begin{tabular*}{\columnwidth}{@{\extracolsep{\fill}}lcc@{}}
		\toprule
		Refinement head & MMD $\downarrow$ & JSD $\downarrow$ \\
		\midrule
		Single convolution & 1.23 & 0.093 \\
		\midrule
		LGR (default) & \textbf{0.94} & 0.079 \\
		LGR (deeper extractor) & 0.95 & \textbf{0.078} \\
		\bottomrule
	\end{tabular*}
\end{minipage}\hfill
\begin{minipage}[t]{\columnwidth}
    \vspace{0pt}
    \captionsetup{type=table}
	\centering
	\small
    \setlength{\tabcolsep}{1.5pt}
	\caption{Loss-weight sensitivity. The first row uses the default LiDAR and ARA weights. Each subsequent row doubles one of these weights while keeping the other weights fixed. MMD is multiplied by $10^4$.}
	\label{tab:supp-loss-weights}
	\renewcommand{\arraystretch}{1.35}
	\begin{tabular}{@{}cccccccc@{}}
		\toprule
		\multicolumn{2}{c}{Loss weights} & \multicolumn{2}{c}{Video quality} & \multicolumn{2}{c}{LiDAR quality} & \multicolumn{2}{c}{CGQ (\%)} \\
		\cmidrule(lr){1-2}\cmidrule(lr){3-4}\cmidrule(lr){5-6}\cmidrule(l){7-8}
		$\lambda_{\mathrm{LiDAR}}$ & $\lambda_{\mathrm{ARA}}$ & FID $\downarrow$ & FVD $\downarrow$ & MMD $\downarrow$ & JSD $\downarrow$ & Global $\uparrow$ & Object $\uparrow$ \\
		\midrule
		1 & 0.01 & \textbf{7.57} & \textbf{30.45} & 0.30 & 0.058 & \textbf{75.68} & 70.16 \\
		2 & 0.01 & 7.68 & 31.12 & \textbf{0.28} & \textbf{0.057} & 75.51 & 70.02 \\
		1 & 0.02 & 7.63 & 30.78 & 0.31 & 0.058 & 75.62 & \textbf{70.28} \\
		\bottomrule
	\end{tabular}
\end{minipage}
\end{table*}

\noindent\textbf{LiDAR self-attention.}
Since LiDAR range maps are sparse and span the full $360^\circ$ azimuth, global self-attention can devote computation to distant regions that contribute little to local geometry. For example, two returns at opposite azimuths lie 64 token columns apart after patch embedding (eight azimuth bins per token). At a horizontal range of 30\,m each, they are 60\,m apart in the ground plane. We therefore restrict attention to azimuthal windows and alternate regular and shifted windows across layers to exchange information between neighboring windows~\cite{swin}. Table~\ref{tab:supp-lidar-self-attention} compares this design with global attention in the LiDAR-only model. Shifted-window attention improves generation quality, reducing MMD from 1.38 to 0.94 and JSD from 0.092 to 0.079. It uses one quarter of the attention FLOPs and reduces forward time per layer from 14.20 to 8.10 ms.

\smallskip
\begin{table*}[!t]
\begin{minipage}[t]{\columnwidth}
    \vspace{0pt}
    \captionsetup{type=table}
	\centering
	\small
	\setlength{\tabcolsep}{3pt}
	\caption{ARA teacher selection. Adaptive chooses the attention direction assigning more mass to the projected object region. Random samples a direction, and Fixed uses video to supervise LiDAR. Global and Object report scene- and object-level CGQ (\%). MMD is multiplied by $10^4$.}
	\label{tab:supp-ara-teacher}
	\normalsize
	\renewcommand{\arraystretch}{1.35}
	\setlength{\tabcolsep}{1.8pt}
	\begin{tabular*}{\columnwidth}{@{\extracolsep{\fill}}lrrrr@{}}
		\toprule
		Teacher & FVD $\downarrow$ & MMD $\downarrow$ & Global $\uparrow$ & Object $\uparrow$ \\
		\midrule
		Adaptive & \textbf{30.45} & \textbf{0.30} & \textbf{75.68} & \textbf{70.16} \\
		Random & 34.29 & 0.41 & 68.13 & 62.41 \\
		Fixed direction & 34.27 & 0.37 & 68.44 & 64.65 \\
		\bottomrule
	\end{tabular*}
\end{minipage}\hfill
\begin{minipage}[t]{\columnwidth}
    \vspace{0pt}
    \captionsetup{type=table}
	\centering
	\footnotesize
	\setlength{\tabcolsep}{2pt}
	\caption{ARA token budget with LGR enabled throughout. No ARA omits the alignment loss. Full ARA selects tokens for every eligible dynamic object, and ARA-fast samples up to 16 eligible tokens per clip. Loss and Step report time spent computing the ARA loss and completing a training step, respectively. Bold marks the better of the two ARA variants in each column.}
	\label{tab:supp-ara-tokens}
	\fontsize{8.4}{11}\selectfont
	\renewcommand{\arraystretch}{1.35}
	\setlength{\tabcolsep}{1.8pt}
	\begin{tabular}{@{}lcccccc@{}}
		\toprule
		\multirow{2}{*}{Variant} & \multicolumn{2}{c}{Quality} & \multicolumn{2}{c}{CGQ (\%)} & \multicolumn{2}{c}{Training time} \\
		\cmidrule(lr){2-3}\cmidrule(lr){4-5}\cmidrule(l){6-7}
		& FVD $\downarrow$ & MMD $\downarrow$ & Global $\uparrow$ & Object $\uparrow$ & Loss (ms) $\downarrow$ & Step (s) $\downarrow$ \\
		\midrule
		No ARA & 32.20 & 0.38 & 70.83 & 64.07 & 0 & 11.84 \\
		\midrule
		Full ARA & \textbf{30.45} & \textbf{0.30} & \textbf{75.68} & \textbf{70.16} & 107.40 & 16.17 \\
		ARA-fast & 30.97 & 0.36 & 71.63 & 68.39 & \textbf{67.58} & \textbf{13.03} \\
		\bottomrule
	\end{tabular}
\end{minipage}
\end{table*}

\noindent\textbf{Cross-modal attention.}
Having examined self-attention within the LiDAR stream, we next test whether video and LiDAR tokens need to interact across all spatial locations. We compare global cross-attention with region-to-region attention that connects each camera view to its corresponding LiDAR region using sensor calibration. Both variants use the same temporal neighborhoods and the full LandV configuration with LGR and ARA enabled. As shown in Table~\ref{tab:supp-cross-modal-attention}, applying cross-attention between corresponding regions improves Global and Object CGQ by 3.05 and 2.87 percentage points, respectively. It also reduces attention FLOPs by 80.1\% and the median forward time per bidirectional layer from 63.24 to 49.61 ms. These results show that sensor calibration can guide cross-modal attention toward better geometric alignment at lower computational cost.

\subsection{Local Geometry Refinement (LGR)}
\label{sec:supp-lgr}
To study LGR's design, we compare three refinement heads in the LiDAR-only model. The baseline uses a single spatial convolution. The default LGR head fuses features from a two-layer local extractor with backbone features, and the deeper variant replaces only the extractor with three narrower convolutional layers. As shown in Table~\ref{tab:supp-lgr-head}, the default LGR head reduces MMD from 1.23 to 0.94 and JSD from 0.093 to 0.079 relative to the single convolution. The deeper head scores similarly (0.95 and 0.078).

\subsection{Loss-weight sensitivity}
\label{sec:supp-loss-weights}

Table~\ref{tab:supp-loss-weights} compares the default loss weights with two variants that double either the LiDAR or the ARA weight while keeping the video weight fixed. Doubling the LiDAR weight from 1 to 2 lowers MMD and JSD but slightly increases FID and FVD. Doubling the ARA weight from 0.01 to 0.02 slightly raises Object CGQ (70.16\% to 70.28\%), while Global CGQ changes little (75.68\% to 75.62\%).

\begin{figure*}[!t]
	\centering
	\includegraphics[width=\textwidth]{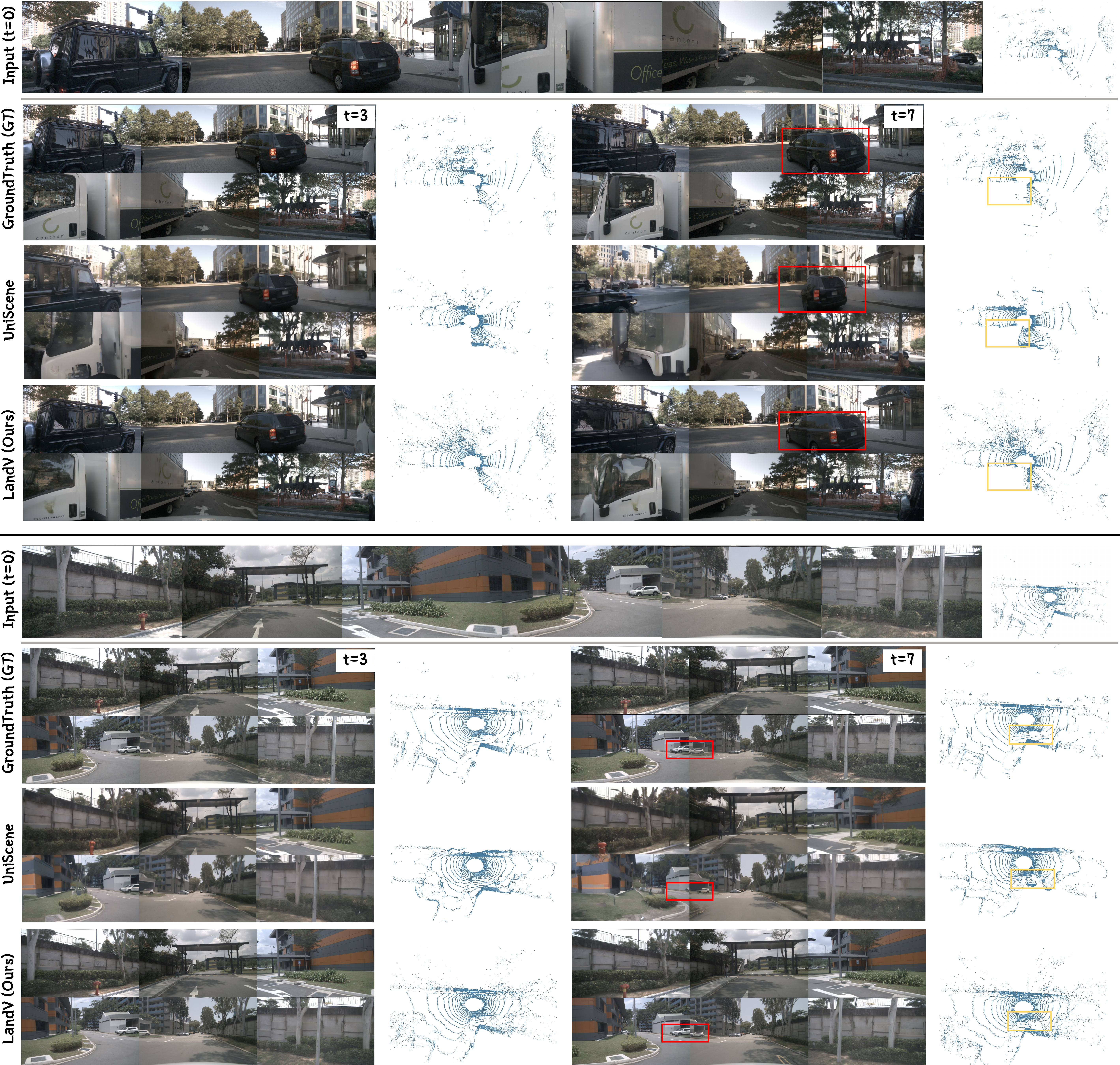}
	\caption{Multi-view video and LiDAR predictions from LandV and UniScene~\cite{uniscene} for two scenes. At $t=7$, red boxes mark corresponding vehicle regions across camera views, and yellow boxes mark the LiDAR regions compared with ground truth.}
	\label{fig:supp-uniscene}
\end{figure*}

\subsection{Adaptive Reciprocal Alignment (ARA)}
\label{sec:supp-ara}
\noindent\textbf{Teacher selection.}
We compare adaptive teacher selection~\cite{ijcai2025p112} with two alternatives, choosing the teacher at random or always using the video stream to supervise the LiDAR stream. Random teacher selection performs worse on all four metrics than the LGR baseline without ARA (Table~4 of the main paper), suggesting that it can introduce unreliable supervision. Using video as the sole teacher yields only modest gains in LiDAR MMD and Object CGQ over this baseline. Adaptive teacher selection achieves the best results across all four metrics (Table~\ref{tab:supp-ara-teacher}).

\smallskip
\noindent\textbf{Token budget.}
Since full ARA selects tokens for every eligible dynamic object throughout the clip, we introduce ARA-fast, which samples up to 16 eligible tokens per clip. Compared with full ARA, ARA-fast reduces ARA loss computation time by 37.1\% and training step time by 19.4\%, while achieving an FVD of 30.97 and an MMD of 0.36 (Table~\ref{tab:supp-ara-tokens}). Relative to LandV with LGR alone (No ARA), ARA-fast improves Global and Object CGQ by 0.80 and 4.32 percentage points, respectively.

\begin{figure*}[!t]
	\centering
	\includegraphics[width=0.95\textwidth]{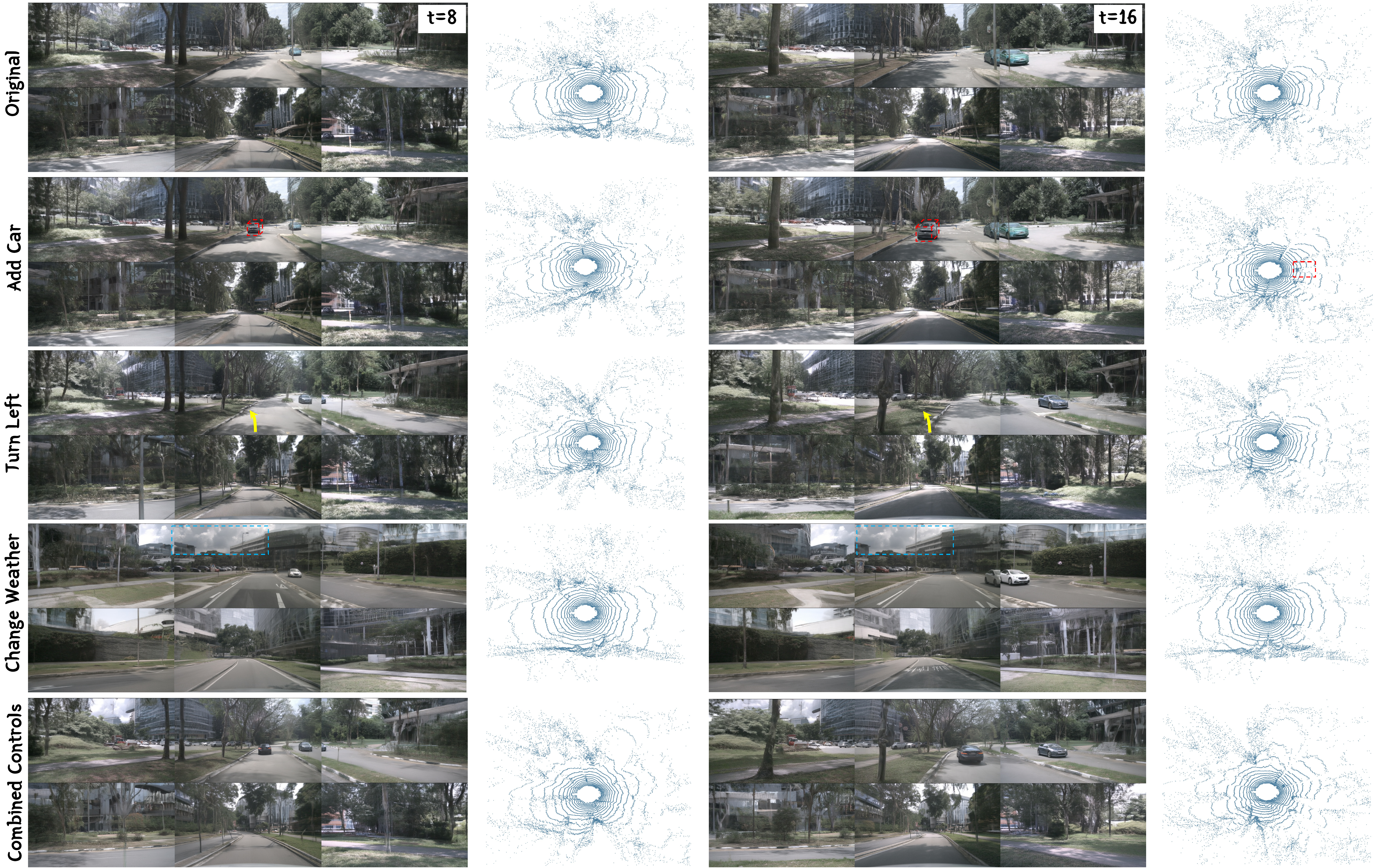}
	\caption{Individual and combined controls at the eighth and sixteenth generated frames. The rows show the original generation, the three individual edits, and their combination. Red boxes mark the added car in the camera and LiDAR outputs, yellow arrows indicate the left turn, and blue boxes mark the cloudy sky.}
	\label{fig:supp-control}
\end{figure*}

\section{Additional Qualitative Experiments}
\label{sec:supp-qualitative}

While the main paper compares LandV with UniScene~\cite{uniscene} over 16 future frames, we further compare both methods at UniScene's native prediction horizon (Section~\ref{sec:supp-qual-uniscene}).
We also evaluate combined scene controls (Section~\ref{sec:supp-qual-control}) and assess the contribution of each modality (Section~\ref{sec:supp-qual-single}).
In all scene-control experiments, we use only HD maps, ego trajectories, 3D boxes, and text prompts, omitting first-frame observations and object appearance features to allow variation in both the environment and vehicle appearance.

\subsection{Comparison with UniScene~\cite{uniscene}}
\label{sec:supp-qual-uniscene}
Given the observations at $t=0$, UniScene~\cite{uniscene} predicts seven future frames in a single call. We restrict the comparison to this window so that both methods predict directly from the same observations. Figure~\ref{fig:supp-uniscene} shows their predictions at $t=3$ and $t=7$ in two scenes. At $t=7$ in the first scene, UniScene shows the rear of the car in one view, but much of its side is missing from the adjacent view (red). Its LiDAR scan also contains a dense, curved cluster where the ground truth has few returns (yellow). LandV preserves the car's side profile across views and more closely matches the sparse distribution of LiDAR returns in the ground truth.

In the second scene at $t=7$, UniScene shows a light-colored parked car beside the building in one view but omits it from the overlapping view (red). In its LiDAR scan, several rounded clusters break up the straight, slanted edge seen in the ground truth (yellow). LandV shows the parked car in both views and preserves the shape of the edge.

\raggedbottom
\subsection{Controllable generation}
\label{sec:supp-qual-control}
To test our LandV as a controllable driving world model, we change the traffic, ego trajectory, and weather separately, then apply the three controls together (Figure~\ref{fig:supp-control}). Specifying a new car by its instance ID and 3D bounding box places it ahead in the front view, with LiDAR returns at the same location (red). Under the left-turn control, the curb shifts across the front view. The cloudy-weather prompt introduces clouds above the buildings. With all three controls, the added car remains visible as the ego vehicle turns, and clouds appear in the front-left view. We further project the LiDAR points into the front camera view to assess whether the two modalities remain aligned under different spatial controls (Figure~\ref{fig:supp-control-additional}). When the ego vehicle turns left, LiDAR returns align with the tree trunk that enters the left foreground. When a car is added, new returns appear at its location in the image. 

\noindent\begin{minipage}{\columnwidth}
	\centering
	\includegraphics[width=0.95\columnwidth]{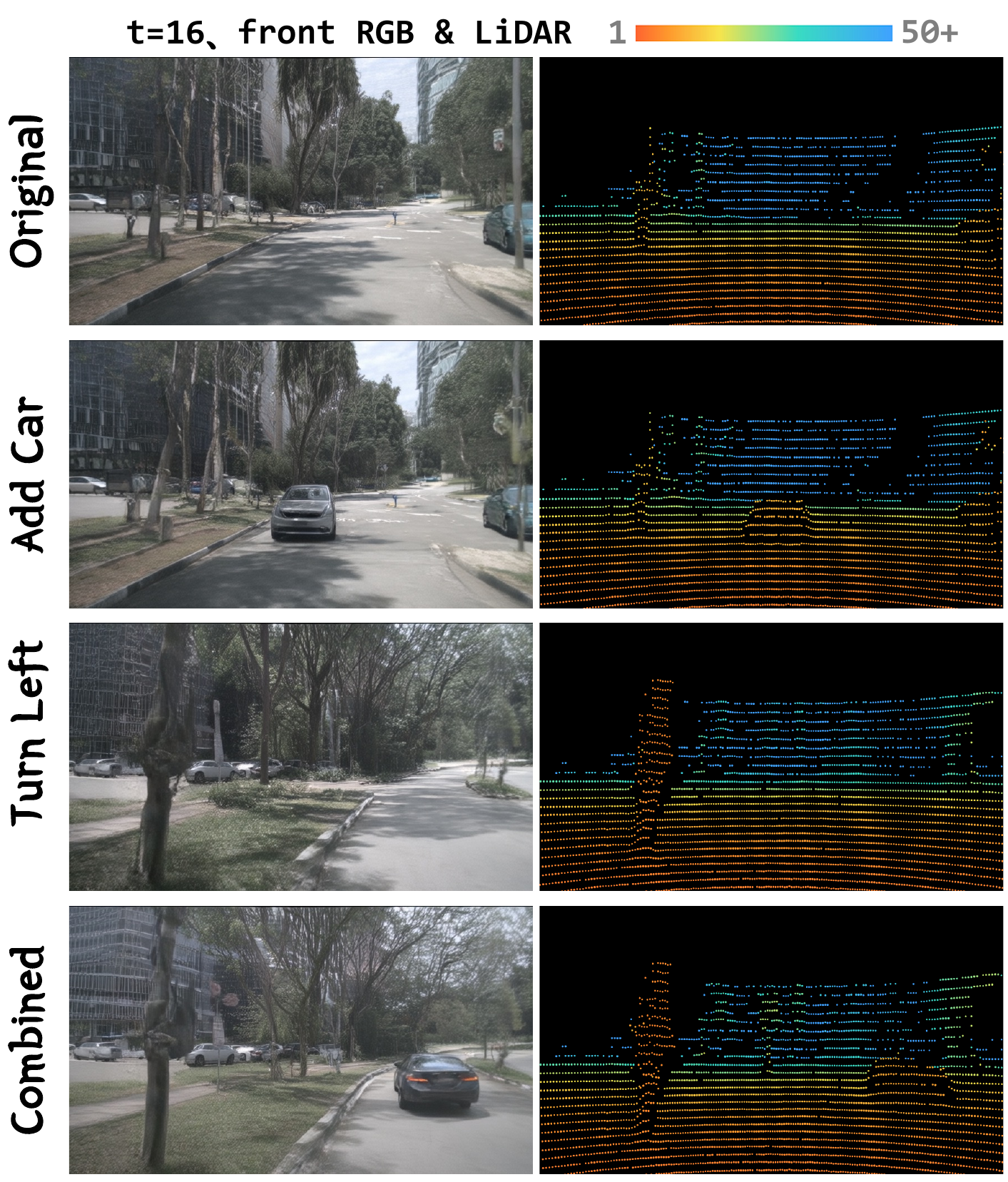}
	\captionof{figure}{Front-view RGB and LiDAR predictions at $t=16$. Both modalities show the added car, the tree brought into view by the left turn, and both together under combined control.}
	\label{fig:supp-control-additional}
\end{minipage}\par

\begin{figure*}[!t]
	\centering
    \captionsetup{skip=4pt}
	\includegraphics[trim=0bp 22bp 0bp 35bp,clip,width=0.95\linewidth]{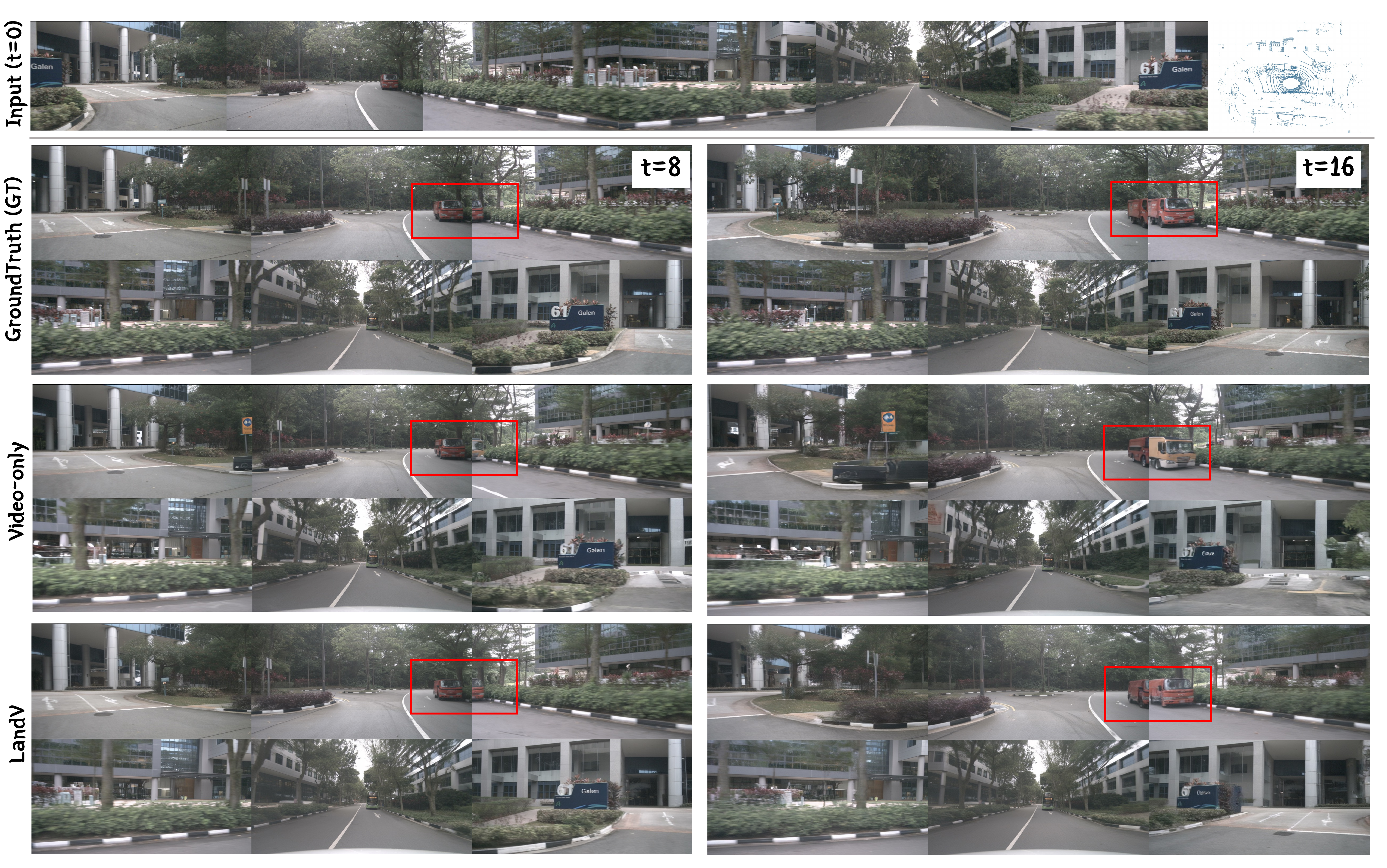}
	\caption{Cross-view appearance of a truck. Red boxes mark its cab in the front and front-right views.}
	\label{fig:supp-video-only}
\par\smallskip
    \captionsetup{type=figure}
	\includegraphics[trim=0bp 13bp 0bp 0bp,clip,width=0.95\linewidth]{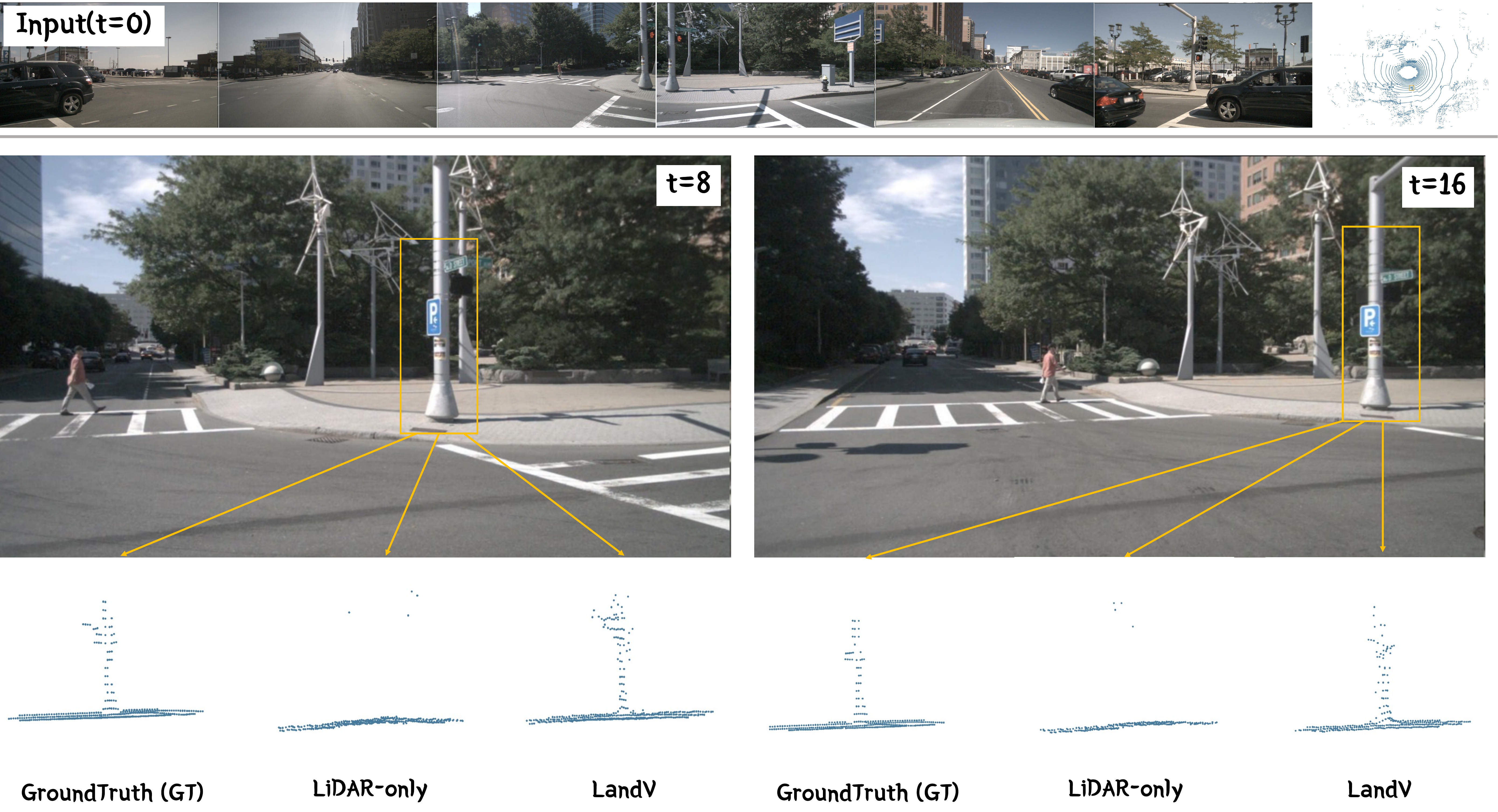}
	\caption{Local LiDAR geometry around a roadside sign. Yellow boxes locate the sign in the camera images. Below each image, the point-cloud crops show ground truth, the LiDAR-only prediction, and the LandV prediction from left to right.}
	\label{fig:supp-lidar-only}
\end{figure*}

\subsection{Comparison with single-modality models}
\label{sec:supp-qual-single}
To examine how video and LiDAR complement each other, we compare LandV with its single-modality variants.
At $t=16$ in Figure~\ref{fig:supp-video-only}, the video-only model renders the truck cab red in the front view and pale yellow in the front-right view, whereas LandV preserves its red appearance across both views, consistent with the ground truth. In Figure~\ref{fig:supp-lidar-only}, the LiDAR-only prediction mainly shows the road surface, while joint generation additionally recovers the upright returns at the roadside sign, as observed in the ground-truth scan.
These examples suggest that LiDAR helps preserve vehicle appearance across cameras, while visual cues help recover small roadside structures in the point cloud.

\flushbottom
\section{OCPR Prompts}
\label{sec:supp-ocpr-prompts}

\subsection{Step 1 Recognition Prompt}
\label{sec:supp-ocpr-prompt-recognition}
\begin{quote}\small
	Inspect this crop of an expected \texttt{\{class\_name\}}.
	
	Can the target be visually recognized as this category? Judge only from visible shape and structure, not the supplied category or background.
	
	Small or partially occluded objects can still be recognizable. However, blurry blobs, distorted structures, or severe generation artifacts are not sufficient.
	
	Return only one JSON object:
	\par\noindent\texttt{\{"recognizable":"yes","reason":"clear"\}}
	
	Use yes/no. For yes, reason is clear. For no, choose from: missing, wrong\_category, blur\_or\_too\_small, occluded\_or\_truncated, ambiguous.
\end{quote}

\subsection{Step 2 Appearance Consistency Prompt}
\label{sec:supp-ocpr-prompt-consistency}
\begin{quote}\small
	Decide whether two crops could show the same \texttt{\{class\_name\}} across \texttt{\{relation\}} in a continuous driving sequence.
	
	The visible side may change completely (front, side, or rear). Compare subtype, 3D shape and proportions, paint color/layout, and structural details visible in both views. Ignore background, crop scale, lighting, and viewpoint.
	
	Answer yes only when visible evidence supports one stable appearance; answer no for a clear contradiction; answer uncertain when an object is missing, unrecognizable, or the views provide too little comparable evidence. Category match alone is not enough for yes, and a different visible side alone is not enough for no.
	
	Return only \texttt{\{"consistent":"yes"\}}, using yes, no, or uncertain.
\end{quote}

For cross-view, cross-time, and joint pairs, we set \texttt{\{relation\}} to \emph{same timestamp, different cameras}, \emph{same camera, separated by 4 frames}, and \emph{different cameras, separated by \texttt{\{dt\_frames\}} frames}, respectively. The value of \texttt{\{dt\_frames\}} is taken from the sample list.

{
    \small
    \bibliographystyle{ieeenat_fullname}
    \bibliography{main}
}